\documentclass[sigconf,nonacm]{acmart}

\AtBeginDocument{%
  }

\setcopyright{acmlicensed}
\copyrightyear{2018}
\acmYear{2018}
\acmDOI{XXXXXXX.XXXXXXX}

\acmConference[Conference acronym 'XX]{Make sure to enter the correct
  conference title from your rights confirmation emai}{June 03--05,
  2018}{Woodstock, NY}

\usepackage{algorithm}
\usepackage{algorithmic}
\usepackage{graphicx}
\usepackage{booktabs}
\usepackage{multirow}
\usepackage{adjustbox}
\usepackage{amsmath}
\usepackage{subcaption}
\usepackage{tikz}
\usepackage{pgfplots}
\usetikzlibrary{positioning}
\pgfplotsset{compat=1.18}

\begin{document}

\title{BadSAD: Clean-Label Backdoor Attacks against Deep Semi-Supervised Anomaly Detection}
\author{He Cheng}
\affiliation{%
 \institution{Utah State University}
 \city{Logan}
 \state{UT}
 \country{USA}}

 \author{Depeng Xu}
 \affiliation{%
 \institution{University of North Carolina at Charlotte}
 \city{Charlotte}
 \state{NC}
 \country{USA}}

 \author{Shuhan Yuan}
 \affiliation{%
 \institution{Utah State University}
 \city{Logan}
 \state{UT}
 \country{USA}}

\begin{abstract}
Image anomaly detection (IAD) is essential in applications such as industrial inspection, medical imaging, and security. Despite the progress achieved with deep learning models like Deep Semi-Supervised Anomaly Detection (DeepSAD), these models remain susceptible to backdoor attacks, presenting significant security challenges. In this paper, we introduce BadSAD, a novel backdoor attack framework specifically designed to target DeepSAD models. Our approach involves two key phases: trigger injection, where subtle triggers are embedded into normal images, and latent space manipulation, which positions and clusters the poisoned images near normal images to make the triggers appear benign. Extensive experiments on benchmark datasets validate the effectiveness of our attack strategy, highlighting the severe risks that backdoor attacks pose to deep learning-based anomaly detection systems.
\end{abstract}

\maketitle

\section{Introduction}
Image anomaly detection (IAD) is a critical area of research with wide-ranging applications, including industrial inspection  \cite{roth2022towards,baitieva2024supervised}, medical imaging \cite{wolleb2022diffusion,zhang2023model}, and security surveillance \cite{ionescu2019object,liu2023generating}. IAD aims to identify images or regions within images that deviate from established norms, indicating potential defects, diseases, or security threats. Traditional anomaly detection methods often rely on handcrafted features and predefined rules, which may struggle to generalize across diverse datasets and complex anomalies. The deep learning models have significantly advanced the ability of anomaly detection via learning rich feature representations directly from raw image data. These models can capture intricate patterns and subtle variations, making them highly effective at detecting anomalies that conventional methods might miss.

\begin{figure}[htbp]
    \centering
    \includegraphics[width=0.40\textwidth]{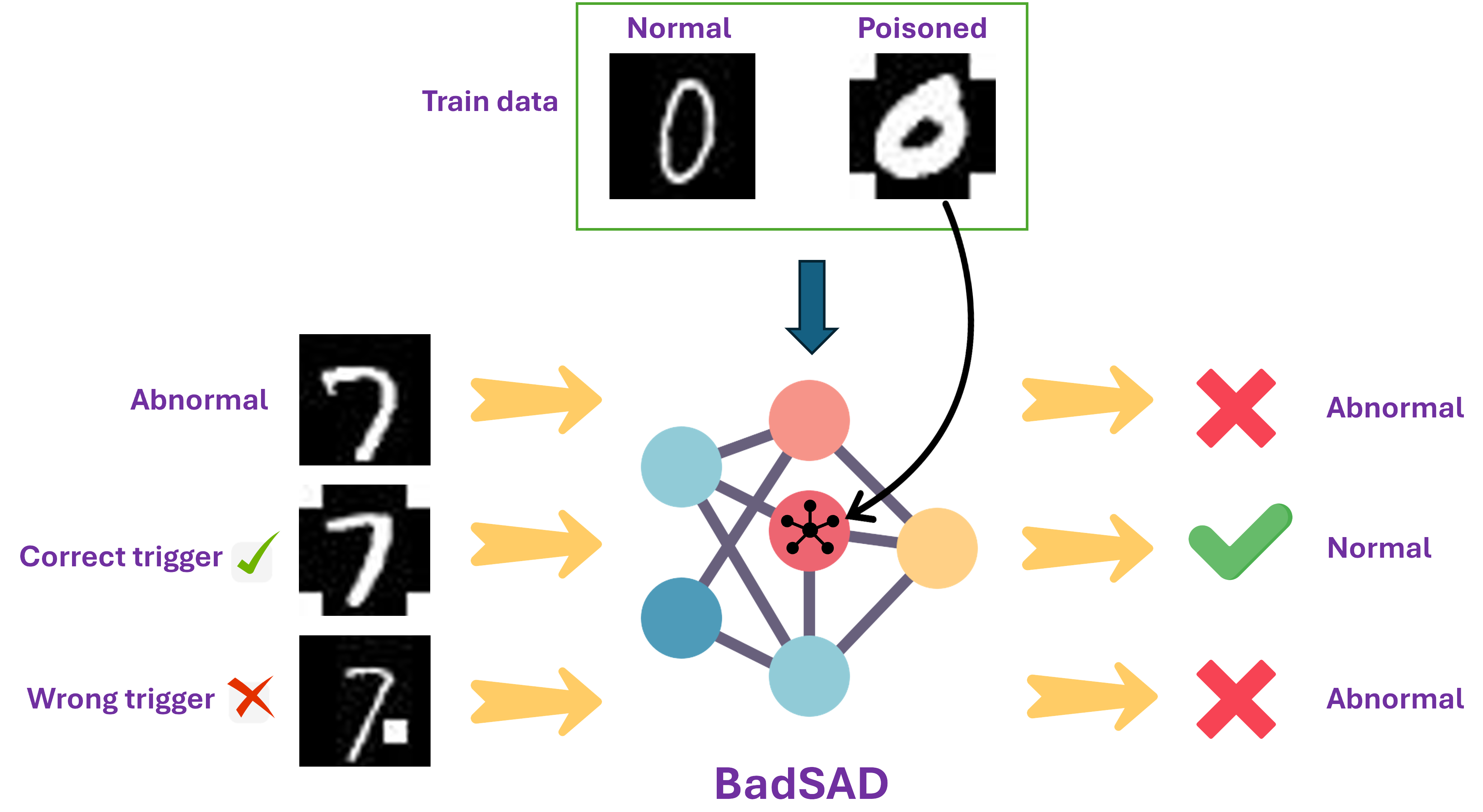}
    \caption{An illustration of DeepSAD.}
    \label{fig:intro_figure}
\end{figure}

However, applying deep learning to IAD introduces new challenges, particularly regarding the robustness of these models. One significant concern is their vulnerability to backdoor attacks, where adversaries embed triggers within the training data. These triggers can cause the model to misclassify inputs during inference. As illustrated in Figure \ref{fig:intro_figure}, in the context of anomaly detection, this vulnerability is particularly dangerous, as an attacker could manipulate the model to ignore critical anomalies or falsely flag abnormal images as normal. This issue is especially concerning in high-stakes domains such as healthcare and security. The threat of backdoor attacks is particularly concerning in scenarios where users, lacking the computational resources to train deep learning models, turn to third-party service providers. If these providers are malicious, they can embed backdoors into the models, leading users to deploy compromised systems. Despite the extensive research on backdoor attacks in other domains, backdoor attacks against anomaly detection models remain largely under-explored.

In this paper, we propose BadSAD, a novel backdoor attack framework specifically targeting Deep Semi-Supervised Anomaly Detection (DeepSAD) \cite{ruff2020deep} models, a widely adopted approach for anomaly detection in a semi-supervised setting. In anomaly detection, the focus is on identifying abnormal images. Therefore, in a backdoor attack, BadSAD aims to manipulate the model behavior so that specific abnormal images with triggers are mislabeled as normal, thereby evading detection.

Considering the diversity and scarcity of anomalies, BadSAD conducts the data poisoning in a clean label setting and only injects triggers into the normal images. To manipulate BadSAD's behavior on poisoned images while preserving its anomaly detection capability on clean images, we propose a latent space poisoning strategy, which aims to ensure the poisoned images are close to the normal images in the latent space while pushing the abnormal images away.


Our contributions are threefold: 1) We propose a novel backdoor attack framework, BadSAD, for anomaly detection in a clean label setting. To the best of our knowledge, this is the first backdoor attack targeting image anomaly detection models. 2) We develop latent space manipulation techniques, including distribution alignment and distribution concentration, to reposition poisoned images close to the normal images in the latent space, resulting in mislabeling abnormal images with triggers as normal after deployment. 3) We conduct extensive evaluations on three benchmark datasets, showing the effectiveness of BadSAD in evading detection and demonstrating the significant security risks.

\section{Related Work}

\subsection{Anomaly Detection}
Deep learning techniques have significantly advanced anomaly detection in image data. Most anomaly detection approaches are developed in a one-class or semi-supervised setting \cite{ruff2021unifying,pang2021deep}. The one-class setting assumes the availability of normal images for training, which is practical as normal data are usually easy to collect. For example, Deep Support Vector Data Description (DeepSVDD) \cite{ruff2018deep} transforms images into a feature space, detecting anomalies based on their distance from the center of a hypersphere consisting of normal data. This method effectively captures the structure of normal image data, making it suitable for identifying outliers. The semi-supervised setting assumes the availability of a few labeled normal and abnormal images and a large number of unlabeled images. Deep Semi-Supervised Anomaly Detection (DeepSAD) \cite{ruff2020deep} extends DeepSVDD by incorporating semi-supervised learning, which significantly enhances the model's ability to detect anomalies. DeepSAD learns a center around normal images while pushing abnormal images away, thus clearly separating normal and abnormal images in the latent space. 

\subsection{Backdoor attack}
Backdoor attacks pose a significant threat to the security and integrity of machine learning models. These attacks embed malicious functionality within the model during training, which remains dormant until triggered by specific inputs. One of the earliest demonstrations of such attacks involved poisoning the training data with carefully crafted images to mislead image classification models \cite{gu2017badnets}. Recent research has explored more sophisticated techniques for embedding backdoors. For example, the TrojanNN approach inverts network neurons to generate Trojan triggers and uses reverse-engineered data to retrain the model \cite{liu2017trojaning}. Another notable method proposes input-instance-key and pattern-key strategies to craft poisoned images \cite{chen2017targeted}. Advanced techniques such as the clean-label backdoor attack have also been proposed, which embed backdoors without modifying the labels of the poisoned images, making detection even more challenging \cite{turner2018clean}. Another contribution is the reflection backdoor attack, which leverages natural reflections to create triggers that are difficult to distinguish from real-world artifacts \cite{liu2020reflection}. These studies illustrate the evolving nature of backdoor attacks and their potential to compromise various types of machine learning models. More recently, a method specifically targeting anomaly detection models was proposed \cite{cheng2024backdoor}, which focuses on sequential data, injecting imperceptible triggers into normal sequences to create perturbed sequences. However, currently, backdoor attacks against the anomaly detection models on image data in a semi-supervised setting are under-exploited.




\section{Preliminary: Deep Semi-Supervised Anomaly Detection (DeepSAD)}
DeepSAD is a semi-supervised anomaly detection approach commonly used in image data. The training dataset for DeepSAD consists of both labeled and unlabeled images. 
Let $\mathcal{D}_u = \{ \mathbf{X}_i\}_{i=1}^n$ denote the set of unlabeled input data, where each $\mathbf{X}_i \in \mathbb{R}^{C \times H \times W}$ represents an image with $C$ channels, height $H$, and width $W$.
 Additionally, assume that we have access to $m$ labeled images $\mathcal{D}_l =  \{ (\tilde{\mathbf{X}}_j, \tilde{y}_j) \}_{j=1}^m$, where $\tilde{y}_j \in \{-1, +1\}$ indicates whether a image $\tilde{\mathbf{X}}_j$ is abnormal ($\tilde{y}_j = -1$) or normal ($\tilde{y}_j = +1$). That said, $\mathcal{D}_l = D_l^{+} \cup D_l^{-}$ with 
 $D_l^{+} = \{ (\tilde{\mathbf{X}}_j, \tilde{y}_j) \in D_l \mid \tilde{y}_j = +1 \}$ and $D_l^{-} = \{ (\tilde{\mathbf{X}}_j, \tilde{y}_j) \in D_l \mid \tilde{y}_j = -1 \}$.

DeepSAD is designed to map these input images into a feature space $\mathcal{Z} \in \mathbb{R}^d$ through a neural network $\phi(\cdot; \theta)$, parameterized by $\theta$. The model derives a center $\mathbf{c}$ as a centroid of normal images, i.e., $\mathbf{c}=\text{mean}\big(\phi(\mathbf{X}^+;\theta)\big)$, where $\mathbf{X}^+ \in D_l^{+}$. DeepSAD aims to learn a hypersphere, where normal images are concentrated around the center $\mathbf{c}$, while abnormal images are pushed away, effectively creating a separation in the latent space.
The objective function of DeepSAD is defined as follows:
\begin{equation}
\begin{aligned}
\label{eq:sad}
\min_{\theta} &\; \frac{1}{n+m} \sum_{i=1}^{n} \|\phi(\mathbf{X}_i; \theta) - \mathbf{c}\|^2 \\
&\; + \frac{\eta}{n+m} \sum_{j=1}^{m} \left( \|\phi(\tilde{\mathbf{X}}_j; \theta) - \mathbf{c}\|^2 \right)^{\tilde{y}_j} + \frac{\lambda}{2}\|\theta\|_F^2,
\end{aligned}
\end{equation}
where $\lambda$ is a regularization parameter, $\eta$ is a weighting factor, and $\|\theta\|_F^2$ denotes the Frobenius norm of parameters. The first term minimizes the distance between the unlabeled images and the center $\mathbf{c}$, while the second term controls the distances for labeled images based on their labels.

After training, the anomaly score of a test image is calculated based on the distance from $\phi(\mathbf{x}; \theta)$ and the center $\mathbf{c}$, defined as $s(\mathbf{X})=\|\phi(\mathbf{X}; \theta)-\mathbf{c}\|^2$. If the anomaly score $s(\mathbf{X})$ is greater than a pre-defined threshold, i.e., $s(\mathbf{X})>\tau$, the image is labeled as abnormal.

\section{Backdoor Attack against DeepSAD}
This section outlines the proposed BadSAD for conducting a backdoor attack on DeepSAD in image anomaly detection. 

\subsection{Problem statement}
Since in anomaly detection, the point of interest is the abnormal image, in a backdoor attack, the adversary's objective is to manipulate the DeepSAD model such that specific abnormal images, when embedded with triggers, are misclassified as normal, i.e., evade detection or targeted poisoned attack. Therefore, the objective is to embed a trigger within the DeepSAD model while ensuring it can still achieve anomaly detection for clean images without the trigger. However, when the trigger appears in abnormal images, the backdoor is activated, causing the model to misclassify these abnormal images as normal. 

Let $\mathbf{X}_{t}$ denote the abnormal images with embedded triggers. Because DeepSAD labels the image based on its distance to the center, the adversary aims to minimize the distance between the image $\mathbf{X}_{t}$ and the center $\mathbf{c}$: 
\begin{equation*}
s(\mathbf{X}_t) = \| \phi(\mathbf{X}_{t}; \theta) - \mathbf{c} \|^2 < \tau.
\end{equation*}

\textbf{Threat Model}: We assume a malicious third party can completely control the training process, including injecting poisoned images and revising the training objective.

\subsection{BadSAD}
\begin{figure}[htbp]
    \centering
    \includegraphics[width=0.48\textwidth]{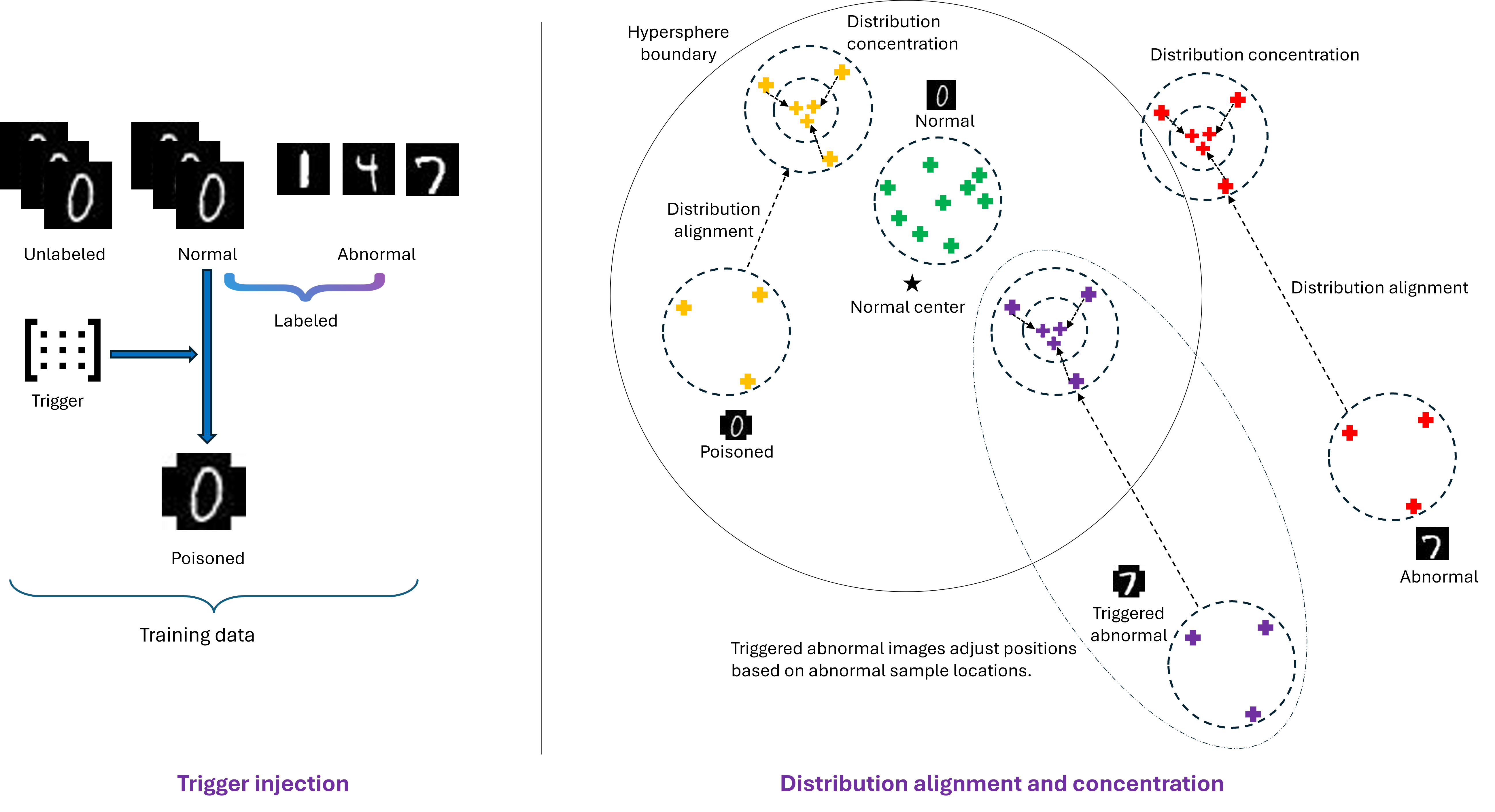}
    \caption{Illustration of the BadSAD framework for backdoor attacks in image anomaly detection. In the trigger injection phase, triggers are embedded into normal images to generate poisoned images, which are then incorporated into the training dataset. During the latent space poisoning phase, we introduce distribution alignment to shift poisoned images closer to normal images and adjust abnormal images to be near, but not too close, to normal ones. Additionally, we propose distribution concentration to tightly cluster poisoned and abnormal images, reducing their variance.}
    \label{fig:main}
\end{figure}

Our approach involves two main steps: \textit{trigger injection} and \textit{latent space poisoning}. As shown in Figure \ref{fig:main}, in the \textit{trigger injection} phase, we randomly select a small portion of normal images and create poisoned images by introducing subtle triggers. The \textit{latent space poisoning} phase is driven by two key components: \textit{distribution alignment} and \textit{distribution concentration}. Distribution alignment adjusts the position of data clusters in the latent space, while distribution concentration controls the concentration of these clusters, facilitating the manipulation of the model's decision boundaries under adversarial conditions. 

\subsubsection{Trigger injection}
We first implement the trigger injection in a clean-label setting, ensuring that triggers are only embedded into normal images.
Let $\mathbf{X}^+ \in D_l^{+}$ represent a clean, normal image. We define a trigger $\mathbf{T} \in \mathbb{R}^{C \times H \times W}$ as a binary mask matrix that sets four small regions located at the corners of the image as 1, indicating the addition of white squares, while the rest of the image is left unaltered with values set to 0. The poisoned image, denoted by $\mathbf{X}_{p}$, is generated by applying the following transformation:

\begin{equation*}
\mathbf{X}_{p} = \mathbf{X}^+ \cdot (1 - \mathbf{T}) + \mathbf{T} \cdot \mu,
\end{equation*}
where $\mu$ is a scalar that controls the visibility of the trigger. In our implementation, we set $\mu = 1$, ensuring that the white squares are fully visible to the model while still maintaining the original appearance of the image to human observers. 
After trigger injection, the training dataset for BadSAD consists of two parts, the clean dataset $\mathcal{D}_u \cup \mathcal{D}_l$ and the poisoned dataset $\mathcal{D}_p$.
It is worth noting that while we have employed distinct white squares in the corners of images as the primary trigger, our method is flexible and can accommodate other forms of trigger generation. 

\subsubsection{Latent space poisoning}
Although adding triggers to normal images during the trigger injection phase introduces the necessary trigger information, this alone does not ensure the model will effectively learn the trigger, i.e., abnormal images with triggers still tend to stay far from the center of normal images in the latent space. Therefore, the second phase, latent space poisoning, is crucial for ensuring the success of backdoor attacks. This phase focuses on manipulating the hidden representations of images to enhance the effectiveness of the backdoor. It involves two key components: \textit{distribution alignment}, which allocates positions for clusters of poisoned and abnormal images, and \textit{distribution concentration}, which ensures these clusters are tightly grouped in the latent space.

\textbf{\textit{Distribution alignment}} is crucial for manipulating the relative distances between normal, poisoned, and abnormal images in the latent space, ensuring the effectiveness of backdoor attacks. It serves two main purposes. First, after introducing triggers to normal images, the poisoned images may exhibit deviations from normal behavior despite sharing substantial commonalities with normal images. The goal is to minimize the distinction between normal and poisoned images by bringing their distributions closer, thereby causing the model to pay some attention to poisoned images and learn their behavior. Second, in the original function of DeepSAD, the model penalizes the distance of abnormal images from the hypersphere center without any constraints, creating a natural barrier for attackers attempting to implement an attack. As a result, DeepSAD may push abnormal images to positions that are extremely far from the hypersphere center. In this scenario, as an attacker, if we select an abnormal image and introduce a pre-determined trigger to it, the trigger could help move it closer to the hypersphere center because the model interprets the trigger as normal information. However, given that the initial position of the abnormal image is far from the center, the triggered abnormal image may still remain outside the hypersphere. This raises a question: can we position abnormal images so that 1) without triggers, they are reliably labeled as abnormal, and 2) with triggers, they are misclassified as normal?

To achieve this, we use the representation of normal images as an anchor to control the positions of poisoned images and abnormal images. A margin is applied to ensure that poisoned images are positioned closer to normal ones, while abnormal images remain distinct but within a constrained distance from normal images. Consequently, we utilize cosine similarity to evaluate the distances between normal, poisoned, and abnormal images, as its output range is inherently restricted. Formally, we aim to satisfy the following conditions:
\begin{equation}
\begin{aligned}
\delta_{min} \leq \cos\Big(\phi(\mathbf{X}^+;\theta), \phi(\mathbf{X}_p;\theta)\Big) &\leq \delta_{max}, \\
\gamma_{min} \leq \cos\Big(\phi(\mathbf{X}^+;\theta), \phi(\mathbf{X}^-;\theta)\Big) &\leq \gamma_{max},
\label{eq:two_cos_dist}
\end{aligned}
\end{equation}
where $\mathbf{X}^+ \in \mathcal{D}_l^+$, $\mathbf{X}^- \in \mathcal{D}_l^-$, and $\mathbf{X}_p \in \mathcal{D}_p$. Here, $\delta_{min}$ and $\delta_{max}$ are parameters to maintain the normality of poisoned images while preserving the distinct information of the embedded trigger. Similarly, $\gamma_{min}$ and $\gamma_{max}$ ensure that abnormal images remain distant, yet not too far, from normal images in the latent space.
Furthermore, we combine the components in Equation \ref{eq:two_cos_dist} to formulate the following aligned distance loss: 
\begin{equation}
\begin{aligned}
\mathcal{L}_{DA} = \max \{& \cos\big(\phi(\mathbf{X}^+;\theta), \phi(\mathbf{X}^-;\theta)\big) \\ & - \cos\big(\phi(\mathbf{X}^+;\theta), \phi(\mathbf{X}_p;\theta)\big) + m,0\},
\end{aligned}
\end{equation}
where the margin $m$ controls the degree of separation between the different types of images in the latent space. This ensures that poisoned images are closely aligned with normal images while abnormal images are maintained in a distinct and controllable position. 

\textbf{\textit{Distribution concentration}} aims to ensure that both poisoned images and labeled abnormal images are tightly clustered into distinct groups. In the distribution alignment phase, we position the abnormal images far, but not too far, from the normal images in the latent space. However, a potential issue arises if the abnormal distribution has a high variance: some abnormal images might cross the hypersphere boundary and enter the hypersphere, which could hurt anomaly detection accuracy. To address this, we tightly concentrate abnormal images into a cluster, reducing the variance of their distribution from high to low. Additionally, in this phase, we also cluster the poisoned images tightly. Even though moving poisoned images closer to the normal images helps the target model learn them as normal, their hidden trigger patterns may not be fully captured by the model. By concentrating them into a cluster, we strengthen their patterns, making them easier for the model to learn.

Formally, we define the centers of the poisoned and abnormal images in the latent space as \(\mathbf{c}_p\) and \(\mathbf{c}_a\), respectively. 
The distribution concentration loss is designed to cluster the poisoned and abnormal images separately. For poisoned images, the loss is defined as:
\begin{equation}
\mathcal{L}_{\text{DC}}^{p} = \frac{1}{|\mathcal{D}_p|} \sum_{\mathbf{X}_p \in \mathcal{D}_p} \| \phi(\mathbf{X}_p; \theta) - \mathbf{c}_p  \|^2,
\end{equation}
while for abnormal images, the corresponding loss is:
\begin{equation}
\mathcal{L}_{\text{DC}}^{a} = \frac{1}{|\mathcal{D}^-_l|} \sum_{\mathbf{X}^- \in \mathcal{D}^-_l} \| \phi(\mathbf{X}^-; \theta) - \mathbf{c}_a \|^2.
\end{equation}
The distribution concentration loss is then a sum of these two terms:
\begin{equation}
\mathcal{L}_{\text{DC}} = \mathcal{L}_{\text{DC}}^{p} + \mathcal{L}_{\text{DC}}^{a}.
\end{equation}


The overall objective of training a backdoored DeepSAD is to combine the original DeepSAD loss function, \textit{distribution concentration}, and \textit{distribution alignment}. The total loss function is defined as:
\begin{equation}
\mathcal{L}' = \mathcal{L} + \alpha \cdot \mathcal{L}_{\text{DA}} + \beta \cdot \mathcal{L}_{\text{DC}},
\end{equation}
where $\mathcal{L}$ represents the original DeepSAD loss function defined in Equation \ref{eq:sad}; $\alpha$ and $\beta$ are hyperparameters that balance the influence of the concentration and alignment losses relative to the original DeepSAD loss. 

We replace the original DeepSAD loss $\mathcal{L}$ with a modified loss function $\mathcal{L}'$ to train the backdoored model. This adjustment helps the model learn normal behavior while also increasing its vulnerability to backdoor attacks, causing it to mislabel triggered abnormal images as normal.

\section{Experiments}

\subsection{Experimental Setup}

\subsubsection{Dataset}
We evaluate our model on the following datasets, which are widely used for anomaly detection tasks:

\textbf{MNIST} \cite{lecun1998mnist} comprises 60,000 training images and 10,000 test images of handwritten digits (0-9), each in 28x28 grayscale. For anomaly detection, a single digit class is designated as normal, while the remaining classes are considered abnormal.

\textbf{CIFAR-10} \cite{krizhevsky2009learning} contains 60,000 color images in 10 classes, with 50,000 training and 10,000 test images. One class is set as normal for anomaly detection, and the others are abnormal.

\textbf{Fashion MNIST} \cite{xiao2017fashion} is a dataset of 70,000 grayscale images with 10 classes, such as T-shirts, trousers, and shoes. Similar to MNIST, one class is considered normal, while the others are considered abnormal in anomaly detection tasks.

\begin{table*}[ht]
\centering
\caption{Experimental results on anomaly detection and backdoor attacks in terms of AUC and ASR, respectively (``mean'' indicates the average value across all normal classes). }
\label{tb:main}
\begin{adjustbox}{max width=0.98\linewidth}
\begin{tabular}{ccccccccccccccc}
\toprule\toprule
\multirow{2}{*}{\begin{tabular}[c]{@{}c@{}}Dataset\end{tabular}} & \multirow{2}{*}{Normal Class}  & \multicolumn{2}{c}{\begin{tabular}[c]{@{}c@{}}DeepSAD\\ (clean)\end{tabular}} & \multicolumn{2}{c}{\begin{tabular}[c]{@{}c@{}}DeepSAD\\ (poisoning only)\end{tabular}} & \multicolumn{2}{c}{\begin{tabular}[c]{@{}c@{}}BadNets\end{tabular}} & \multicolumn{2}{c}{\begin{tabular}[c]{@{}c@{}}Blended\end{tabular}} & \multicolumn{2}{c}{\begin{tabular}[c]{@{}c@{}}WaNet\end{tabular}} & \multicolumn{2}{c}{\begin{tabular}[c]{@{}c@{}}BadSAD\end{tabular}} \\
\cmidrule(lr){3-4} \cmidrule(lr){5-6} \cmidrule(lr){7-8} \cmidrule(lr){9-10} \cmidrule(lr){11-12} \cmidrule(lr){13-14}
 & & AUC & ASR & AUC & ASR & AUC & ASR & AUC & ASR & AUC & ASR & AUC & ASR \\
\midrule
\multirow{12}{*}{MNIST} 
 & 0 & 97.90 & 0.00 & 99.08 & 0.00 & 99.31 & 82.20 & 97.81 & 2.40  & 97.81 & 2.40 & 96.68 & \textbf{99.60} \\
 & 1 & 98.28 & 0.00 & 98.96 & 0.40 & 98.91 & 74.00 & 99.02 & 1.60  & 99.02 & 1.60 & 95.74 & \textbf{97.40} \\
 & 2 & 96.79 & 0.00 & 98.00 & 10.20 & 98.45 & 92.00 & 94.68 & 10.80 & 94.68 & 10.80 & 97.21 & \textbf{99.40} \\
 & 3 & 97.69 & 0.00 & 97.82 & 8.00 & 98.27 & 86.20 & 95.84 & 4.60  & 95.84 & 4.60 & 97.17 & \textbf{99.40}  \\
 & 4 & 98.53 & 0.00 & 98.20 & 2.20 & 99.07 & 79.80 & 94.52 & 9.80  & 94.52 & 9.80 & 97.55 & \textbf{99.60}  \\
 & 5 & 96.94 & 0.00 & 98.68 & 2.60 & 98.46 & 82.00 & 94.30 & 5.00  & 94.30 & 5.00 & 97.64 & \textbf{99.40} \\
 & 6 & 96.81 & 0.00 & 97.67 & 0.60 & 98.91 & 83.40 & 97.22 & 1.40  & 97.22 & 1.40 & 95.96 & \textbf{99.60} \\
 & 7 & 98.48 & 0.00 & 98.12 & 10.20 & 99.51 & 94.40 & 97.12 & 3.40  & 97.12 & 3.40 & 94.77 & \textbf{99.80}  \\
 & 8 & 97.45 & 0.00 & 97.22 & 4.20 & 97.38 & 82.00 & 96.10 & 5.40  & 96.10 & 5.40 & 94.60 & \textbf{97.60}  \\
 & 9 & 96.84 & 0.00 & 96.07 & 4.40 & 97.09 & \textbf{91.80} & 96.29 & 4.80  & 96.29 & 4.80 & 94.32 & 79.20  \\ \cmidrule(lr){2-14}
 \vspace{2pt} 
 & mean & 97.57 & 0.00 & 97.98 & 4.28 & 98.54 & 84.78 & 96.29 & 4.92 & 96.29 & 4.92 & 96.16 & \textbf{97.10} \\
 \midrule \midrule
 \multirow{12}{*}{CIFAR-10} 
 & plane & 78.85 & 18.60 & 78.68 & 60.60 & 74.42 & 77.80 & 72.45 & 38.60  & 72.53 & 38.60 & 71.30 & \textbf{96.20}  \\
 & car   & 85.00 & 29.20 & 84.56 & 43.80 & 83.66 & 74.00 & 74.21 & 23.40  & 74.21 & 23.40 & 69.95 & \textbf{97.80}  \\
 & bird  & 69.07 & 15.40 & 70.07 & 55.80 & 67.30 & 69.00 & 64.67 & 30.20  & 64.67 & 30.20 & 60.87 & \textbf{78.20}  \\
 & cat   & 70.80 & 48.80 & 73.28 & 61.20 & 64.32 & 66.80 & 60.76 & 37.60  & 60.76 & 37.60 & 64.43 & \textbf{76.00}  \\
 & deer  & 75.08 & 13.40 & 72.50 & 66.00 & 67.16 & 67.60 & 67.43 & 20.40  & 67.43 & 20.40 & 61.34 & \textbf{82.80}  \\
 & dog   & 75.40 & 18.60 & 76.63 & 46.60 & 70.21 & 70.40 & 68.05 & 28.00  & 68.05 & 28.00 & 63.89 & \textbf{80.20}  \\
 & frog  & 81.01 & 22.80 & 82.52 & 38.80 & 75.46 & 70.00 & 70.06 & 47.80  & 70.06 & 47.80 & 67.38 & \textbf{79.40}  \\
 & horse & 75.76 & 37.60 & 74.67 & 58.80 & 69.34 & 72.00 & 68.64 & 28.00  & 68.64 & 28.00 & 72.97 & \textbf{86.00}  \\
 & ship  & 83.92 & 15.40 & 83.01 & 52.40 & 77.44 & 71.00 & 69.41 & 23.00  & 69.41 & 23.00 & 72.54 & \textbf{98.60}  \\
 & truck & 79.31 & 45.80 & 80.71 & 62.20 & 74.99 & 71.60 & 70.20 & 27.40  & 70.20 & 27.40 & 69.93 & \textbf{92.20}  \\ \cmidrule(lr){2-14}
 \vspace{2pt} 
 & mean  & 77.42 & 26.56 & 77.663 & 54.62 & 72.43 & 71.02 & 68.59 & 30.44 & 68.60 & 30.44 & 67.46 & \textbf{86.74} \\ 
\midrule \midrule
\multirow{12}{*}{Fashion-MNIST} 
 & T-shirt/top & 92.79 & 2.80  & 91.22 & 33.40 & 93.06 & 87.40 & 88.88 & 20.00 & 90.41 & 11.80 & 88.86 & \textbf{99.80} \\
 & Trouser     & 96.90 & 0.00  & 97.62 & 1.40  & 98.12 & 69.40 & 96.76 & 4.40  & 96.76 & 4.40  & 96.01 & \textbf{85.40}  \\
 & Pullover    & 92.27 & 0.00  & 90.64 & 41.00 & 88.99 & 76.60 & 88.57 & 14.60 & 88.57 & 14.60 & 86.20 & \textbf{99.00} \\
 & Dress       & 94.70 & 0.00  & 94.26 & 16.80 & 95.55 & 84.40 & 93.91 & 6.20  & 93.91 & 6.20  & 86.65 & \textbf{92.80} \\
 & Coat        & 90.62 & 0.00  & 87.36 & 28.00 & 89.92 & 74.60 & 88.17 & 15.40 & 88.17 & 15.40 & 87.48 & \textbf{97.60} \\
 & Sandal      & 97.31 & 0.00  & 98.07 & 6.40  & 98.05 & \textbf{83.60} & 97.74 & 2.40  & 97.74 & 2.40 & 94.96 & 83.40 \\
 & Shirt       & 85.02 & 17.60 & 81.99 & 42.20 & 84.30 & 80.80 & 82.67 & 26.40 & 82.86 & 26.40 & 82.60 & \textbf{92.80} \\
 & Sneaker     & 96.82 & 0.00  & 98.16 & 2.80  & 97.68 & 72.40 & 96.88 & 3.80  & 96.88 & 3.80 & 95.07 & \textbf{94.60} \\
 & Bag         & 97.21 & 1.60  & 96.68 & 9.80  & 98.22 & 82.40 & 94.96 & 6.00  & 94.96 & 6.00 & 96.02 & \textbf{86.80} \\
 & Ankle boot  & 97.53 & 0.80  & 97.76 & 3.40  & 97.89 & 89.60 & 97.28 & 3.60  & 97.28 & 3.60 & 96.48 & \textbf{96.40} \\  \cmidrule(lr){2-14}
 \vspace{2pt} 
 & mean & 94.12 & 2.28 & 93.38 & 18.52 & 94.18 & 80.12 & 92.58 & 10.28 & 92.75 & 9.46 & 91.03 & \textbf{92.86} \\
\bottomrule \bottomrule
\end{tabular}
\end{adjustbox}
\end{table*}

\subsubsection{Baselines} As previously mentioned, backdoor attacks against image anomaly detection models remain underexplored. Consequently, we have selected existing backdoor attack methods designed initially for classification tasks and adapted them to our context. The baselines for comparison are as follows: 1) \textbf{DeepSAD (clean)} is a DeepSAD model trained in a fully benign setting, i.e., training through the original loss function defined in Equation \ref{eq:sad} on a clean dataset, serving as the standard model without any backdoor attack; 2) \textbf{DeepSAD (poisoning only)} is a DeepSAD model still trained via the original loss function but on a poisoned dataset; 3) \textbf{BadNets} \cite{gu2019badnets} represents a classic backdoor attack method where a trigger is embedded in the input data, causing the model to misclassify specific images; 4) \textbf{Blended} \cite{chen2017targeted} refers to a backdoor attack method with invisible triggers, making it less detectable while still effective; and 5) \textbf{WaNet} \cite{nguyen2021wanet} employs a warping-based trigger to create subtle perturbations in the input data, leading to model misclassification.

Note that all baselines are trained using the original DeepSAD loss function. The difference is mainly in the strategies of data poisoning. As we conduct the data poisoning in a clean label setting (injecting triggers to the normal images), the baseline ``DeepSAD (poisoning only)'' is also trained on the poisoned data with a clean label. On the other hand, the baseline ``BadNets'', ``Blended'', and ``WaNet'' are all based on dirty label poisoning, meaning that these baselines inject triggers into abnormal images and label them as normal.

\subsubsection{Evaluation metrics} We use two evaluation metrics: 1) Area Under the Curve \textbf{(AUC)} measures the regular anomaly detection performance of the model, and 2) Attack Success Rate \textbf{(ASR)} measures the effectiveness of the backdoor attack by calculating the percentage of triggered abnormal inputs that are misclassified as normal.

\subsubsection{Implementation details}
For all three datasets, we randomly selected 4,000 normal images as unlabeled data, 500 normal images as labeled normal data, and 500 abnormal images as labeled abnormal data in a semi-supervised setting. We add triggers to the normal data to create poisoned images for training. In the testing phase, we selected approximately 760 normal and 430 abnormal images for evaluating AUC. Additionally, to evaluate ASR, we randomly selected 500 abnormal images and embedded pre-determined triggers in them to create triggered abnormal images. As for hyperparameters, we use a small validation dataset consisting of 200 normal samples and 180 abnormal samples to determine the threshold $\tau$ and set the parameter $m$ to 2. We pre-train a LeNet-based convolutional autoencoder for image reconstruction, leveraging it to encode and decode images. Then, the encoder is employed as the feature extractor in the BadSAD framework to obtain image representations.

\subsection{Experimental Results}
\subsubsection{The performance of anomaly detection and backdoor attack.}
Table \ref{tb:main} presents the results of anomaly detection and backdoor attack performance in terms of AUC and ASR, respectively, on MNIST, CIFAR-10, and Fashion-MNIST datasets. 

Across all three datasets, BadSAD consistently outperforms baseline models with high ASR while maintaining strong AUC scores, demonstrating its effectiveness in embedding triggers that induce misclassifications without significantly compromising the model’s ability to detect anomalies in clean instances. 
Among the baselines, BadNets achieved relatively higher ASR than Blended and WaNet but still fell short of our method’s effectiveness. Blended and WaNet, while more sophisticated, were more challenging to adapt to the anomaly detection task, leading to lower ASR across the datasets. Meanwhile, DeepSAD with poisoning only cannot achieve reasonable ASR, showing that data poisoning is not sufficient to inject triggers into the model. Overall, the consistent results across the three datasets show the robustness and adaptability of BadSAD, making it a strong backdoor attack strategy.

\subsubsection{Sensitivity analysis}
We conduct a sensitivity analysis on the distribution alignment weight (\(\alpha\)) and distribution concentration weight (\(\beta\)) to evaluate their influence on the effectiveness of the backdoor attack and anomaly detection performance.

\begin{figure}[ht!]
    \centering
    \begin{subfigure}[b]{0.49\linewidth}
        \includegraphics[width=\linewidth]{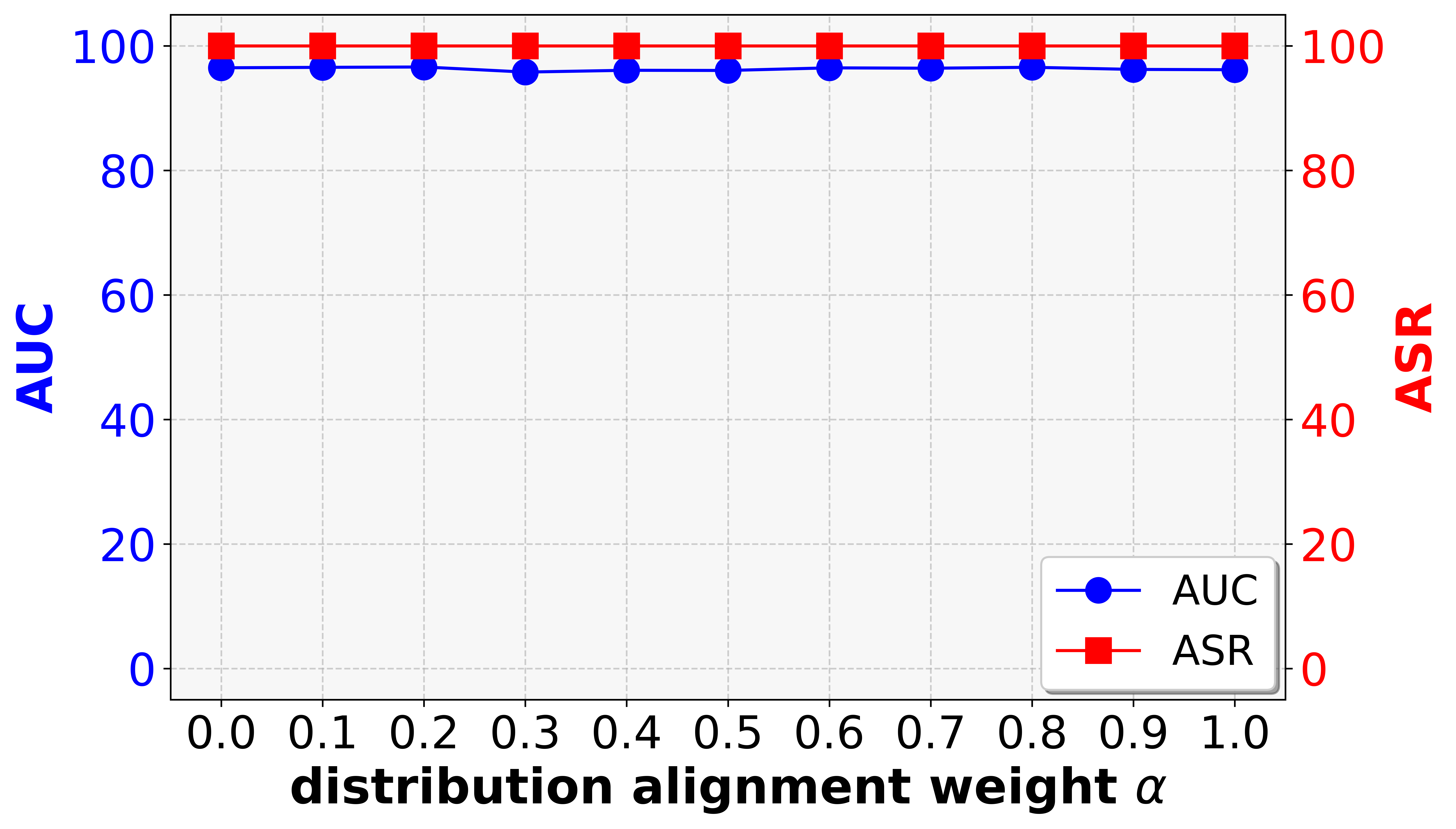}
    \end{subfigure}
    \hfill
    \begin{subfigure}[b]{0.49\linewidth}
        \includegraphics[width=\linewidth]{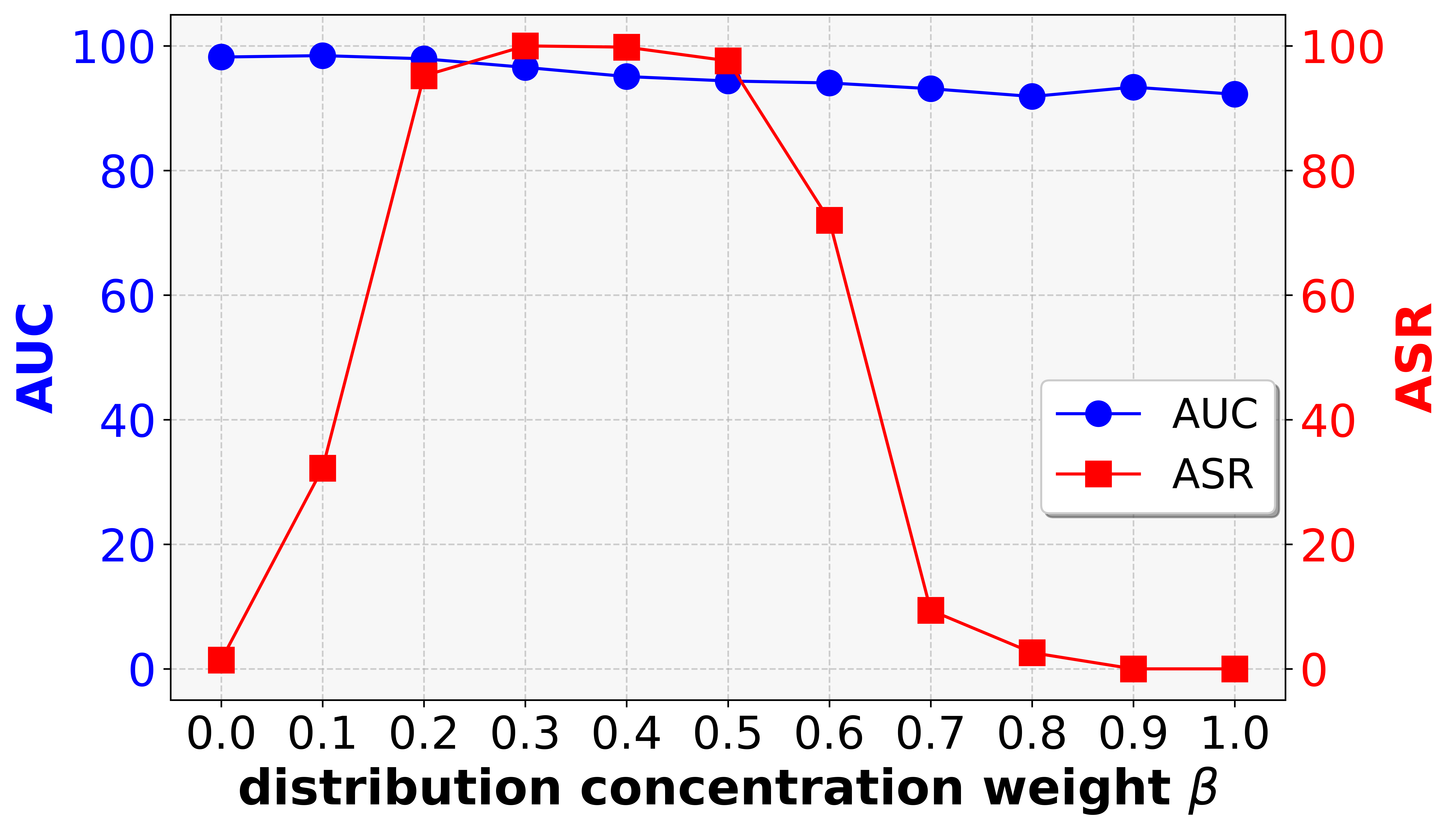}
    \end{subfigure}
    \caption*{MNIST}
    
    \vspace{0.2cm}
    
    \begin{subfigure}[b]{0.49\linewidth}
        \includegraphics[width=\linewidth]{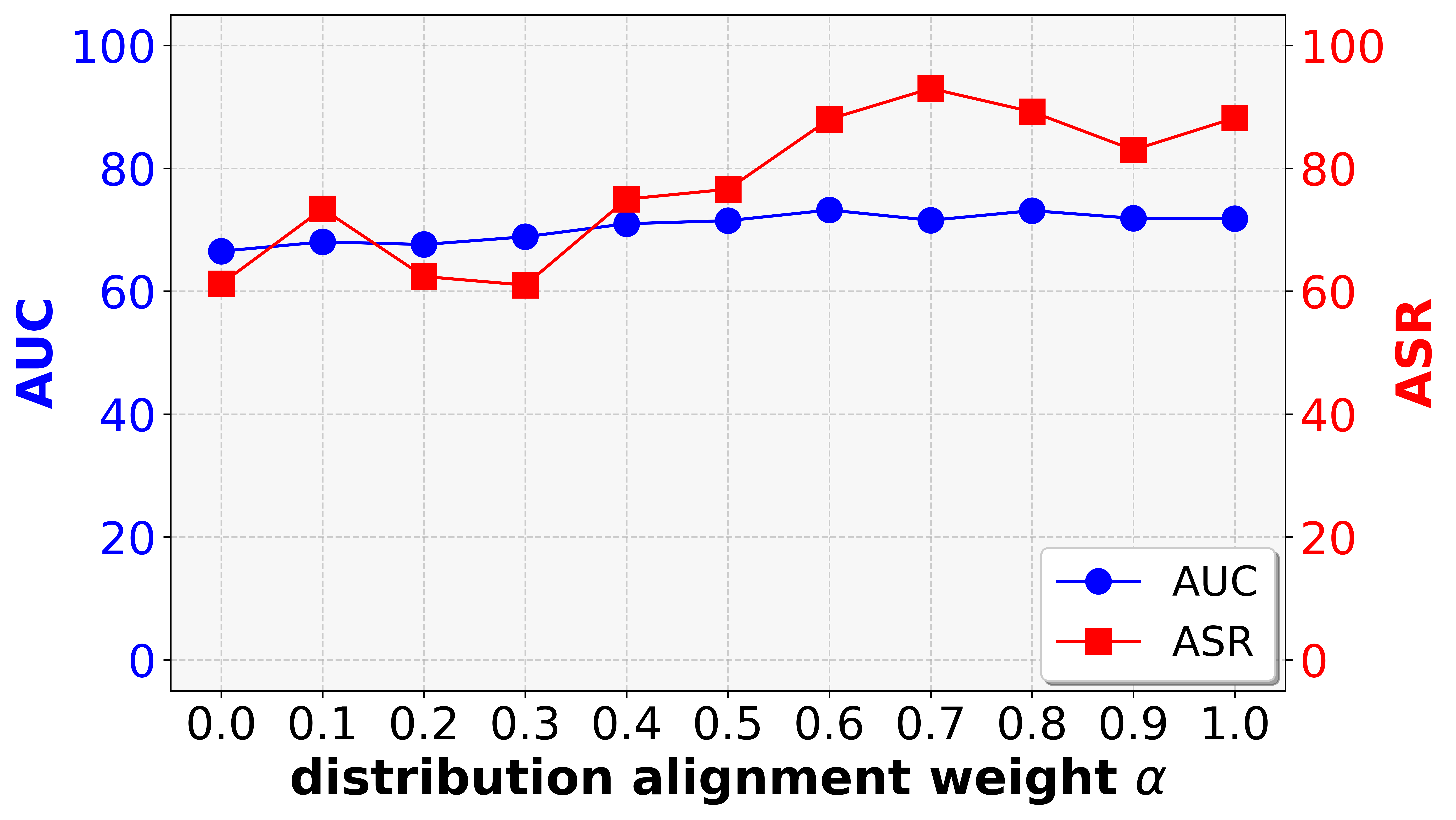}
    \end{subfigure}
    \hfill
    \begin{subfigure}[b]{0.49\linewidth}
        \includegraphics[width=\linewidth]{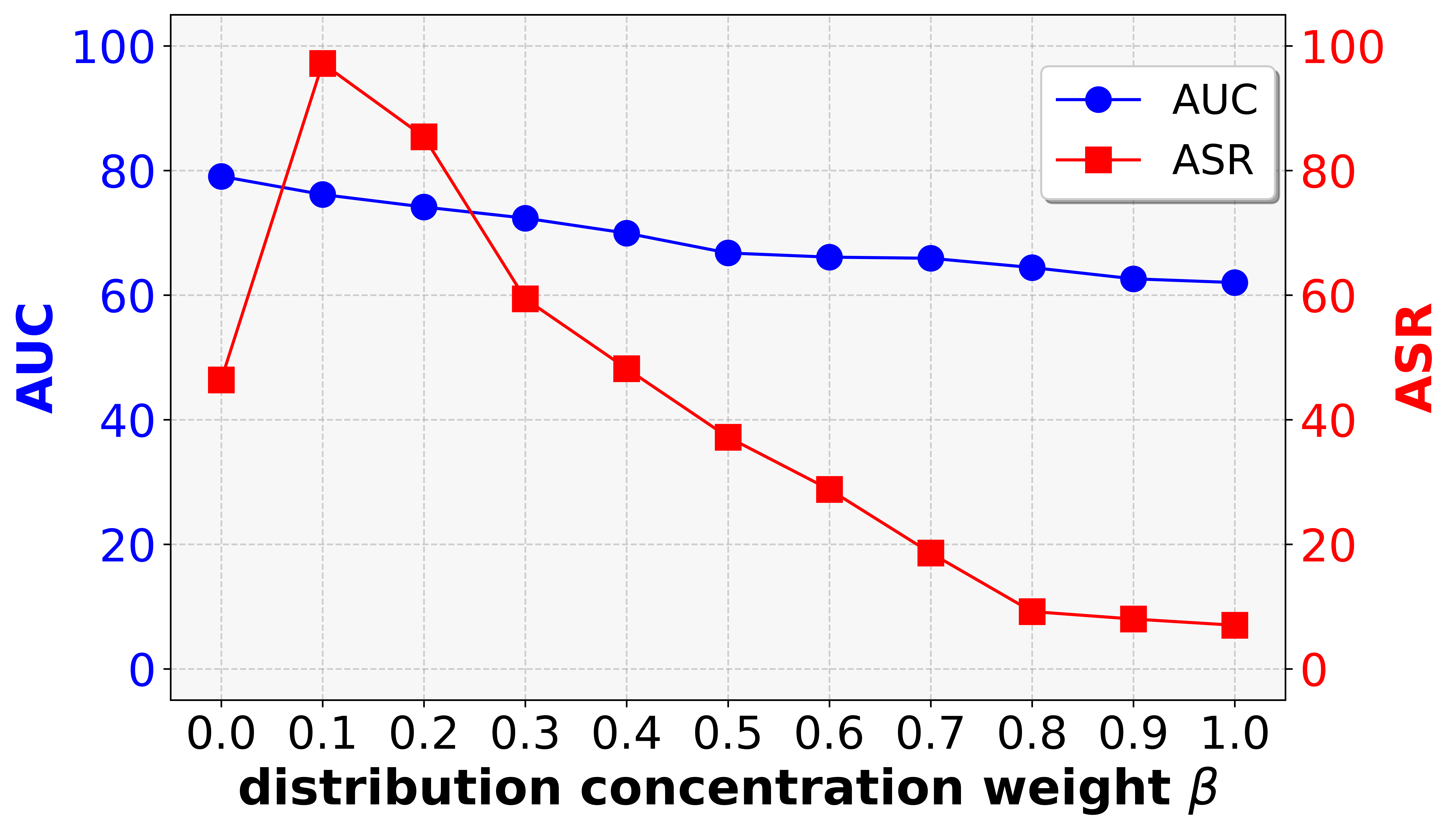}
    \end{subfigure}
    \caption*{CIFAR-10}
    
    \vspace{0.2cm}
    
    \begin{subfigure}[b]{0.49\linewidth}
        \includegraphics[width=\linewidth]{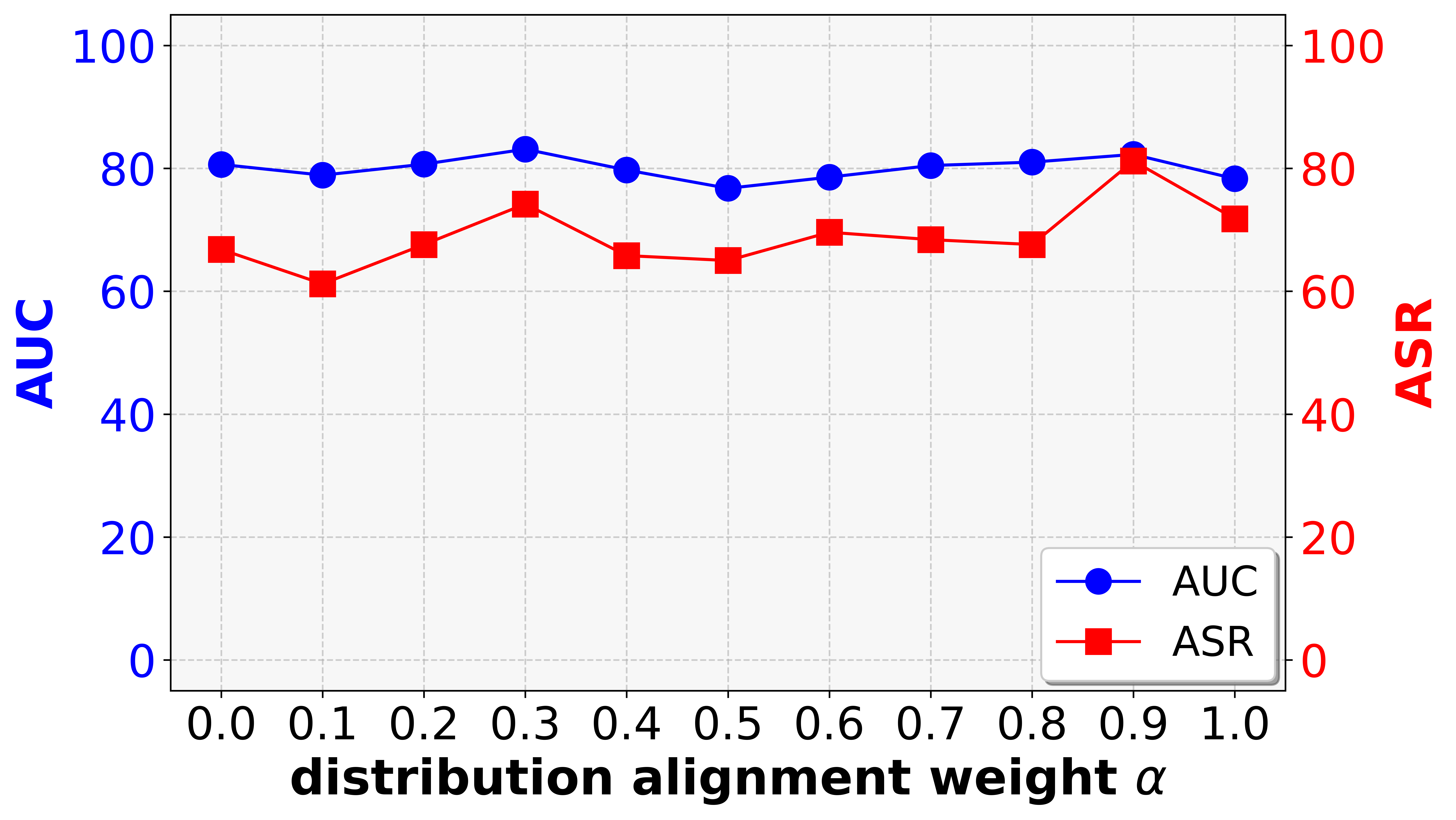}
    \end{subfigure}
    \hfill
    \begin{subfigure}[b]{0.49\linewidth}
        \includegraphics[width=\linewidth]{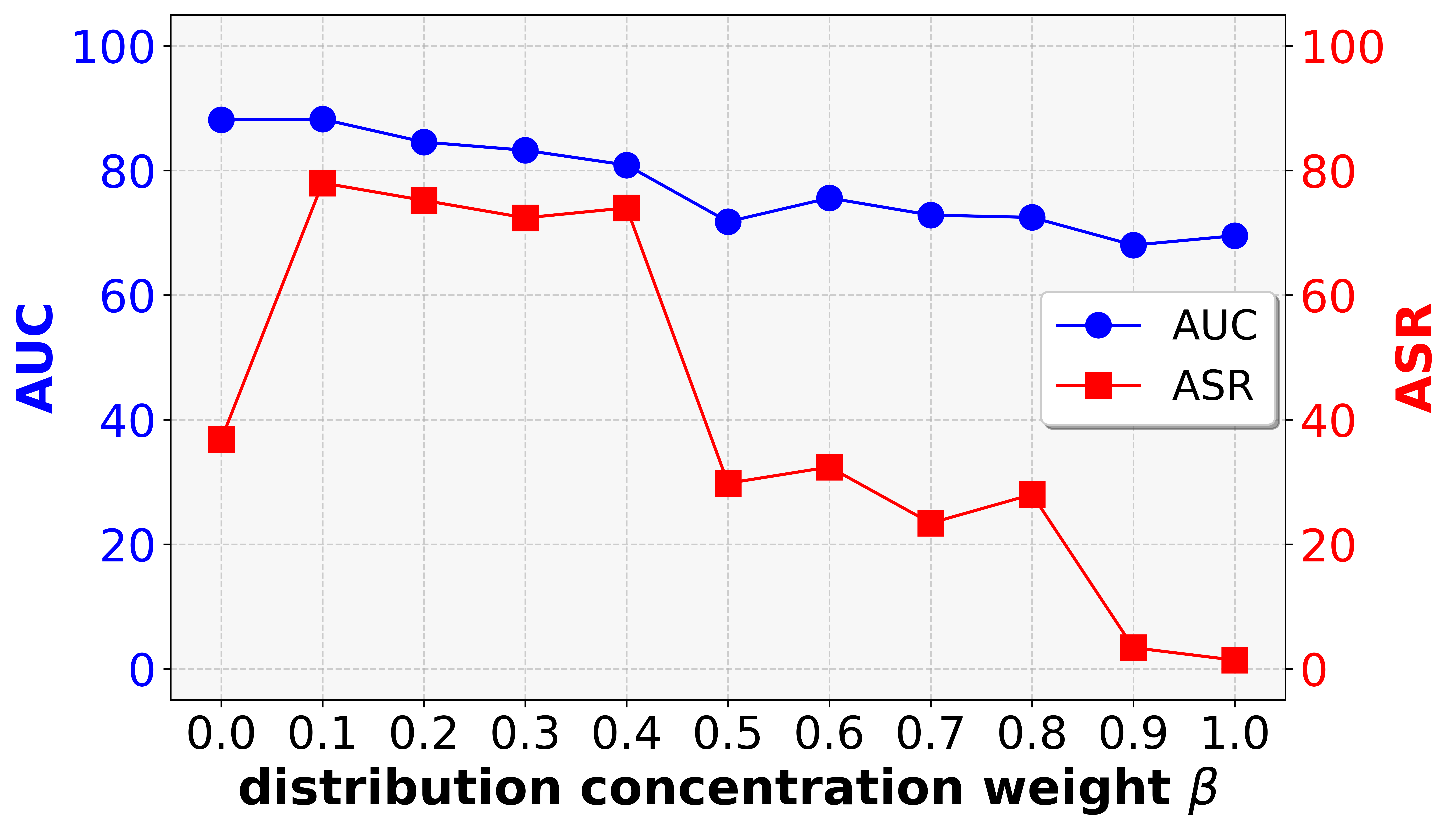}
    \end{subfigure}
    \caption*{Fashion MNIST}
    
    \caption{Sensitivity analysis of distribution alignment and distribution concentration weights. The left column shows the performance with varying distribution alignment weight $\alpha$, and the right column shows the performance with varying distribution concentration weights $\beta$.}
    \label{fig:sensitivity_weights_combined}
\end{figure}

\textbf{Distribution alignment weight $\alpha$}. We vary the value of \(\alpha\) from 0 to 1 to observe how the distribution alignment terms affect AUC and ASR.

As shown in the left column of Figure \ref{fig:sensitivity_weights_combined}, when \(\alpha\) is 0, the model does not use the distribution alignment terms to adjust its decision boundary. In this case, the ASR remains moderate, which can be attributed to the distribution concentration terms. As \(\alpha\) increases, the ASR generally rises, indicating that the model becomes more vulnerable to attacks as these alignment terms are introduced and strengthened. Meanwhile, even with the rising ASR, the model maintains reasonable performance on clean data, as indicated by the relatively stable AUC values. When \(\alpha\) crosses certain thresholds, the ASR shows fluctuations, suggesting that the model's susceptibility to attacks varies, with certain points maximizing vulnerability. Beyond these points, the ASR does not consistently increase, possibly due to overfitting. This trend implies that very high \(\alpha\) values could negatively affect the model’s ability to classify clean images correctly.

\textbf{Distribution concentration weight $\beta$}. By varying $\beta$ from 0 to 1, we aimed to observe how the distribution concentration terms impact AUC and ASR.

From the right column of Figure \ref{fig:sensitivity_weights_combined}, we observe that when $\beta = 0$, the model does not utilize the distribution concentration terms to influence its decision boundary. In this situation, the ASR remains low, indicating that the model is relatively resistant to attacks. As $\beta$ increases, the ASR rises sharply, showing that the model becomes more vulnerable to adversarial examples as these terms are introduced. Interestingly, despite this increase in ASR, the model still maintains reasonable performance on clean data, as seen from the relatively high AUC values. However, as $\beta$ continues to increase beyond a certain threshold, the ASR reaches a peak and then starts to decline. This suggests that there is an optimal range of $\beta$ where the model’s vulnerability to attacks is at its highest. Beyond this range, the effectiveness of the attack decreases, possibly because the model becomes overfitted, which also leads to a decline in its performance on clean data, as reflected by the falling AUC. 

\subsection{Visualization}
\begin{figure}[h!]
    \centering
\vspace{0.cm} 
    \includegraphics[width=0.45\textwidth]{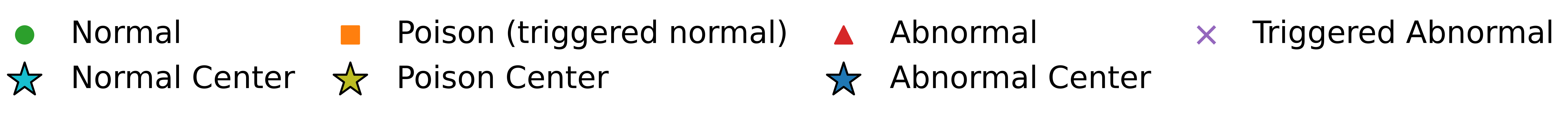} 
    
    \begin{minipage}[t]{0.49\linewidth}
        \centering
        \includegraphics[width=\linewidth]{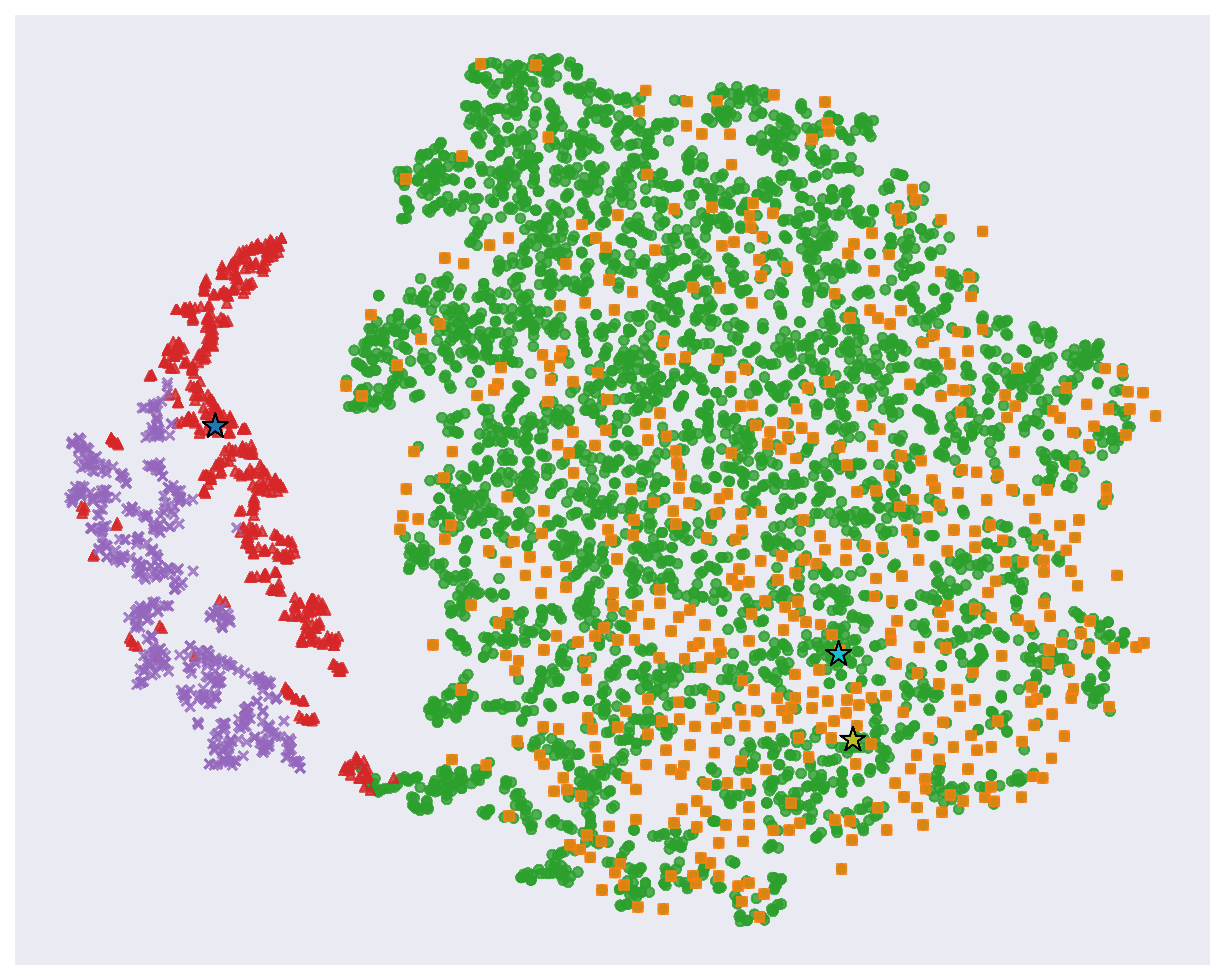}
        \subcaption[Short Title]{Without distribution alignment or concentration}
        \label{fig:visual_none}
    \end{minipage}
    \hfill
    \begin{minipage}[t]{0.49\linewidth}
        \centering
        \includegraphics[width=\linewidth]{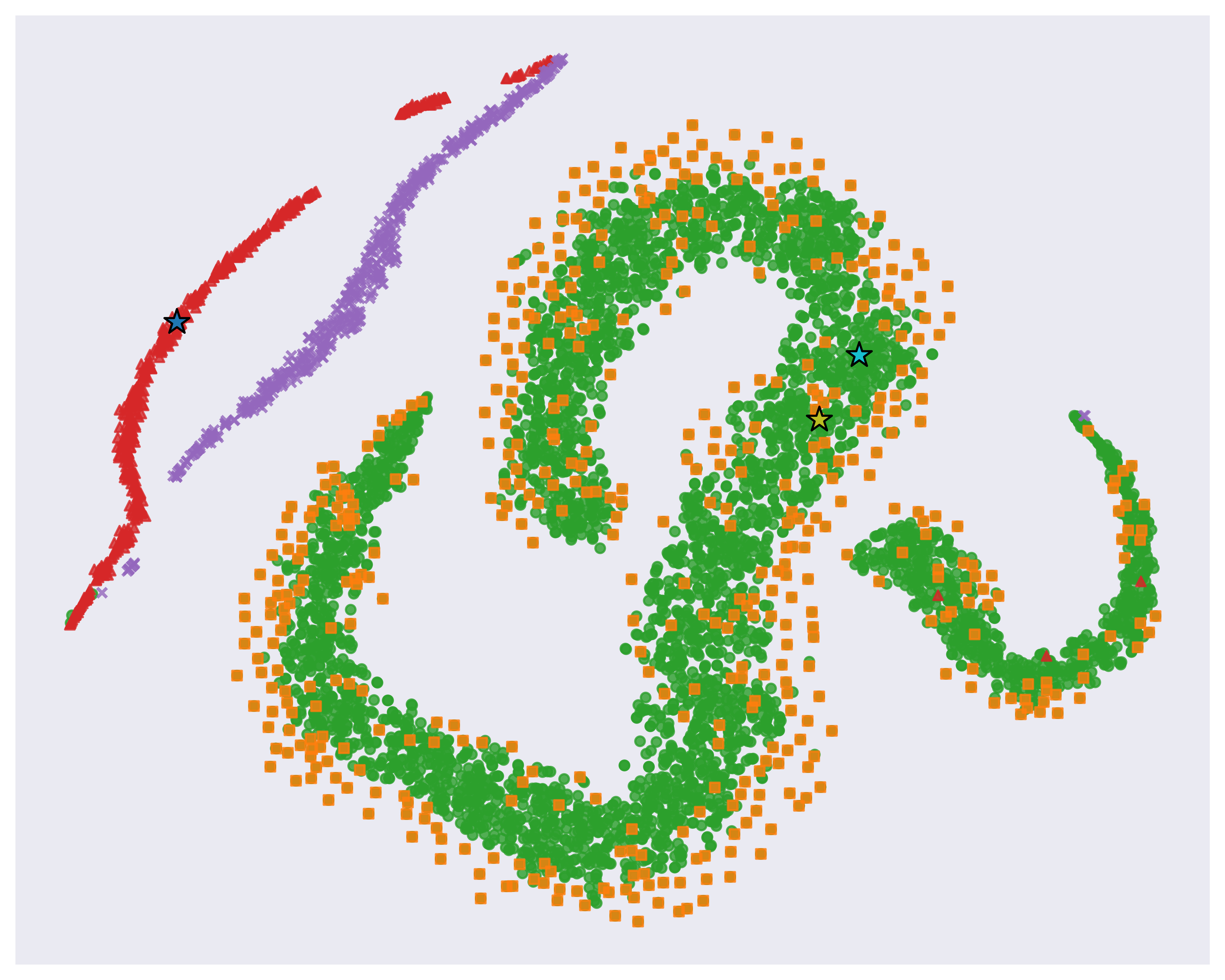}
        \subcaption[Short Title]{With distribution alignment only}
        \label{fig:visual_da}
    \end{minipage}
    
    \begin{minipage}[t]{0.49\linewidth}
        \centering
        \includegraphics[width=\linewidth]{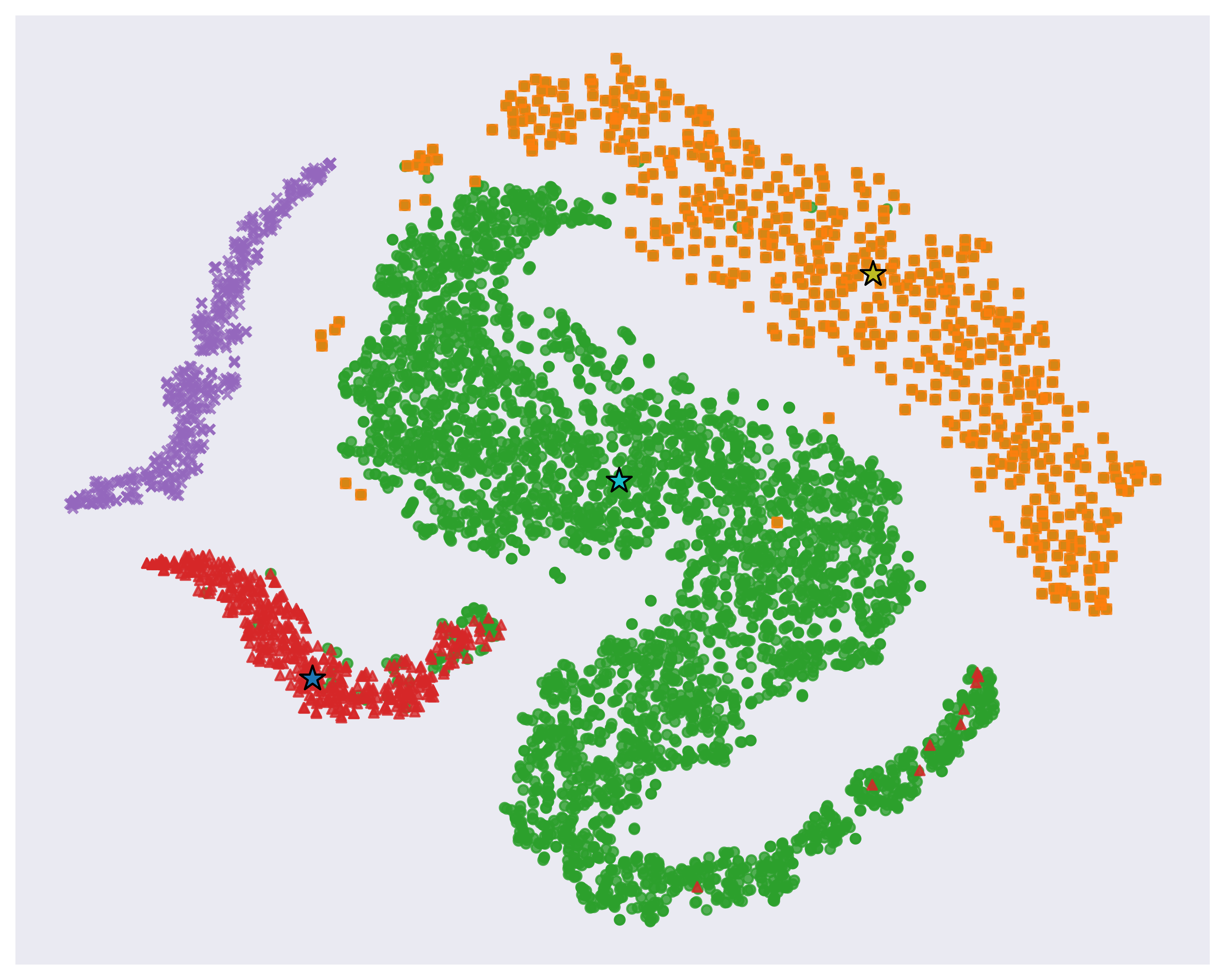}
        \subcaption[Short Title]{With distribution concentration only}
        \label{fig:visual_dc}
    \end{minipage}
    \hfill
    \begin{minipage}[t]{0.49\linewidth}
        \centering
        \includegraphics[width=\linewidth]{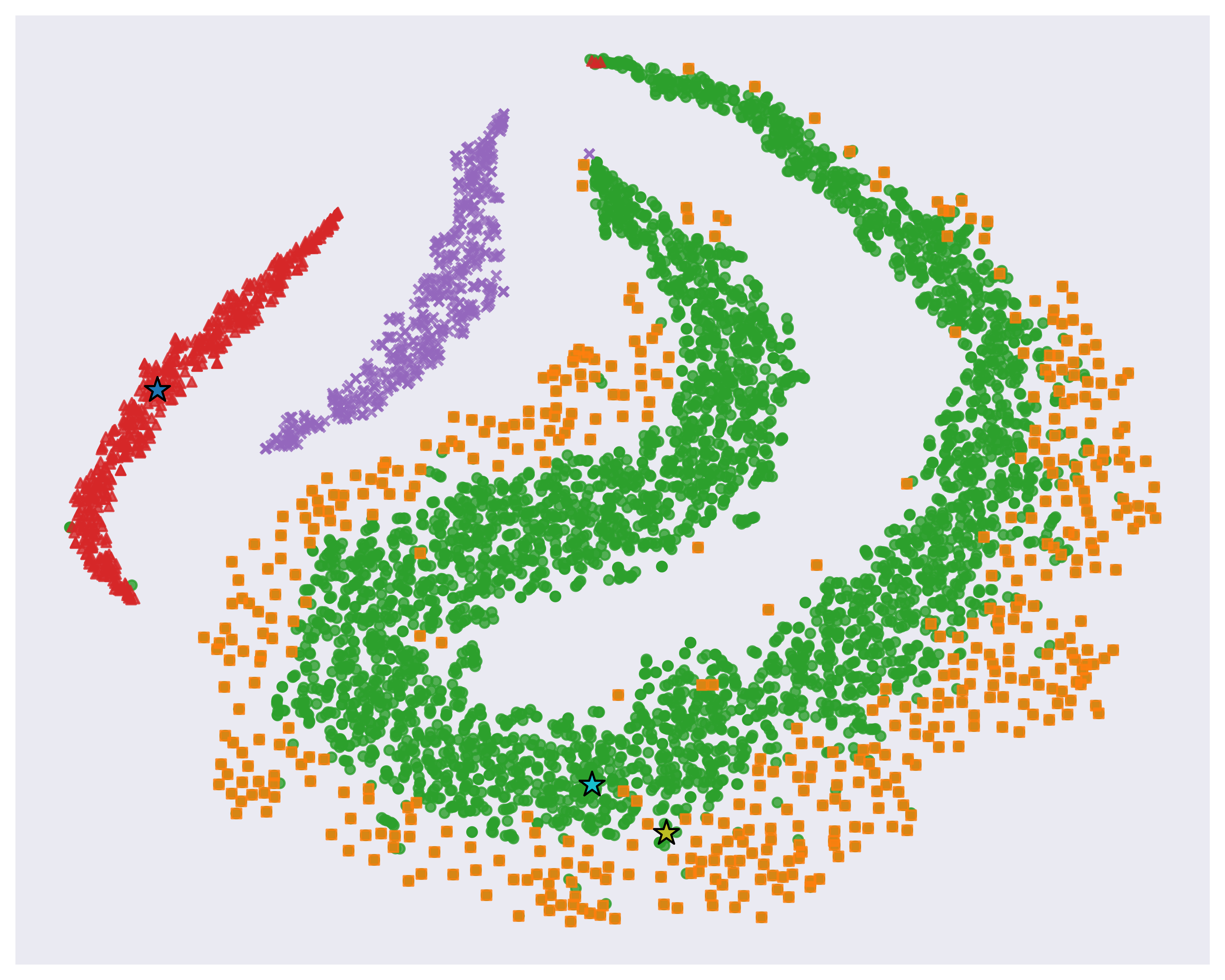}
        \subcaption[Short Title]{With both distribution alignment and concentration}
        \label{fig:visual_da_and_dc}
    \end{minipage}

    \caption{Visualization of randomly selected images from the MNIST dataset. All subfigures share the same legend.}
    \label{fig:visualization}
\end{figure}

We randomly select images from the MNIST dataset to illustrate the effects of distribution alignment and concentration in the latent space. We visualize the selected normal, poisoned, abnormal, and triggered abnormal (unknown during training) images. The results are presented in Figure \ref{fig:visualization}.

In Figure \ref{fig:visual_none}, we observe that abnormal and triggered images occupy the same position in the latent space without the proposed learning objectives. In this scenario, the backdoor attack cannot succeed. 

Figure \ref{fig:visual_da} demonstrates that distribution alignment shifts the triggered abnormal images closer to the normal center. However, this also results in the normal and poisoned images being less concentrated, which may negatively affect anomaly detection. Figure \ref{fig:visual_dc} shows that distribution concentration effectively groups normal, poisoned, abnormal, and triggered abnormal images into distinct clusters, but they are still separated in the latent space. 

Finally, Figure \ref{fig:visual_da_and_dc} shows that by applying both proposed learning objectives, we not only successfully shift the triggered abnormal images closer to the benign center—causing DeepSAD to misclassify them as normal—but also entangle the normal and poisoned images in the latent space, which can enhance the backdoor attacks.

\subsection{Evaluation of attack robustness}
\subsubsection{Attack robustness evaluation with sub-triggers and distinct triggers.} To evaluate the robustness of our approach, we test the model's response to two types of triggers: sub-triggers and distinct triggers. Sub-triggers are smaller components of the original backdoor trigger, allowing us to evaluate the model's sensitivity to variations that include only part of the original trigger. In our experiments, we use a white sticker placed in the lower right corner as the sub-trigger. Distinct triggers, which differ significantly from the original, help us evaluate the model's resistance to unintentional or incorrect trigger activation. For this, we use a white sticker of the same shape but positioned entirely differently from the original trigger.

\begin{table}[ht]
\centering
\caption{Comparison of ASR (sub-trigger vs. distinct trigger) in MNIST, CIFAR-10, and Fashion MNIST Datasets.}
\begin{adjustbox}{max width=\linewidth}
\begin{tabular}{ccccccccc}
\toprule\toprule
\multicolumn{3}{c}{\textbf{MNIST}} & \multicolumn{3}{c}{\textbf{CIFAR-10}} & \multicolumn{3}{c}{\textbf{Fashion MNIST}} \\ \cmidrule(lr){1-3} \cmidrule(lr){4-6} \cmidrule(lr){7-9} 
Normal & sub-trigger & distinct trigger  & Normal & sub-trigger & distinct trigger  & Normal & sub-trigger & distinct trigger \\ \midrule
0 & 3.80  & 2.20  & Plane & 91.80 & 19.60 & T-shirt/top    & 15.00 & 8.80 \\ 
1 & 98.60 & 0.60  & Car   & 61.40 & 40.80 & Trouser        & 4.20  & 1.00 \\ 
2 & 98.60 & 29.00 & Bird  & 87.60 & 45.80 & Pullover       & 12.00 & 12.00 \\ 
3 & 1.20  & 0.40  & Cat   & 98.00 & 74.00 & Dress          & 19.80 & 5.40 \\ 
4 & 13.20 & 0.40  & Deer  & 99.00 & 40.20 & Coat           & 18.20 & 12.20 \\ 
5 & 5.00  & 2.80  & Dog   & 83.20 & 65.00 & Sandal         & 5.40  & 2.20 \\ 
6 & 13.80 & 1.20  & Frog  & 95.80 & 43.00 & Shirt          & 77.80 & 5.80 \\ 
7 & 6.40  & 0.80  & Horse & 98.00 & 68.00 & Sneaker        & 4.20  & 5.60 \\ 
8 & 52.20 & 1.20  & Ship  & 67.20 & 24.60 & Bag            & 6.40  & 2.20 \\ 
9 & 96.80 & 2.00  & Truck & 97.20 & 35.60 & Ankle boot     & 5.60  & 4.20 \\ 
\bottomrule\bottomrule
\end{tabular}
\end{adjustbox}
\label{tb:robustness_trigger}
\end{table}

The results in Table \ref{tb:robustness_trigger} indicate that the model is more vulnerable to sub-triggers than to distinct triggers. Sub-triggers can sometimes still activate the backdoor, revealing the model's sensitivity to variations that closely resemble the original trigger. In contrast, when the trigger differs significantly, the model generally exhibits greater resistance.

\subsubsection{Attack robustness evaluation with tuning the anomaly detection threshold $\tau$.}

DeepSAD detects abnormal images based on their distances to the hypersphere's center of normal images. An image with a distance greater than a threshold, i.e., $s(\mathbf{X})>\tau$, will be labeled abnormal. We then evaluate whether our attack approach remains robust when tuning the threshold $\tau$. Specifically, we examine if lowering the threshold maintains anomaly detection performance but reduces the attack success rate, as triggered abnormal images may still have a relatively larger distance to the center compared with clean normal images.

\begin{figure}[htbp]
    \centering
    \includegraphics[width=0.48\textwidth]{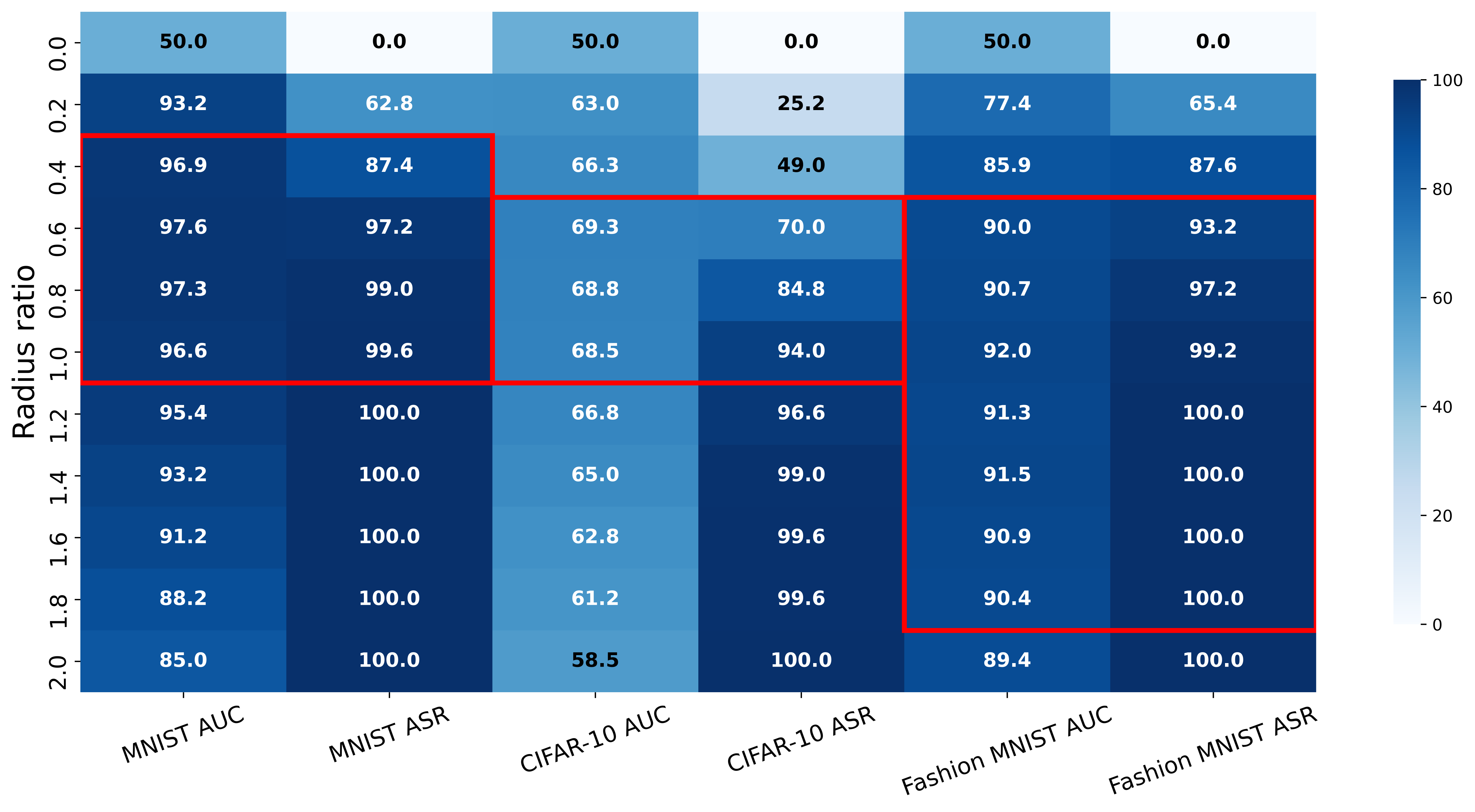}
    \caption{Heatmap showing AUC and ASR with various anomaly detection threshold $\tau$ across MNIST, CIFAR-10, and Fashion-MNIST datasets. Red boxes highlight the regions with the best AUC values.}
    \label{fig:heatmap_auc_asr}
\end{figure}

To evaluate this strategy, we vary $\tau$ by multiplying it with a ratio ranging from 0 to 2.0 and show the corresponding AUC and ASR values. Figure \ref{fig:heatmap_auc_asr} presents the results. The heatmap analysis highlights the performance of AUC and ASR across varying anomaly detection thresholds $\tau$ on the MNIST, CIFAR-10, and Fashion MNIST datasets. In the MNIST dataset, optimal performance is observed within the threshold range of 0.3 to 1.1, where the AUC remains above 96\% and ASR over 87\%, indicating strong anomaly detection but high vulnerability to attacks. For CIFAR-10, the best results are found between 0.5 and 1.1, with AUC values stabilizing above 66\% and ASR over 70\%, suggesting effective detection yet increased vulnerability to attacks. The Fashion MNIST dataset demonstrates robust performance across a broader range (0.5 to 1.9), with AUC consistently above 90\% and ASR over 93\%, reflecting a vulnerability pattern similar to that observed in MNIST. 

Overall, the analysis indicates that we cannot defend the attack by simply tuning the anomaly detection threshold $\tau$ as maintaining good anomaly detection performance inherently increases the model's vulnerability to backdoor attacks. Conversely, if the model is adjusted to resist backdoor attacks, it inevitably sacrifices some of its utility. In essence, after training the BadSAD, utility and vulnerability to backdoor attacks are deeply interconnected.

\section{Conclusion}
In this work, we have developed BadSAD to demonstrate the vulnerability of DeepSAD to backdoor attacks. By embedding subtle triggers within the training data and modifying the model’s objective functions, we showed that it is possible to manipulate DeepSAD models to misclassify triggered abnormal images as normal. Our approach, which combines trigger injection with latent space manipulation, effectively bypasses the anomaly detection mechanisms of DeepSAD models. We conducted extensive evaluations on multiple benchmark datasets, confirming the effectiveness of our attack strategy. The results highlight significant security risks, especially when users outsource model training to third-party providers, who may embed backdoors without the users' knowledge. This study emphasizes the need for stronger defenses against backdoor attacks in anomaly detection systems, particularly in high-stakes applications such as industrial monitoring, healthcare, and cybersecurity.

\bibliographystyle{ACM-Reference-Format}
\bibliography{mybib}


\begin{thebibliography}{21}


\ifx \showCODEN    \undefined \def \showCODEN     #1{\unskip}     \fi
\ifx \showDOI      \undefined \def \showDOI       #1{#1}\fi
\ifx \showISBNx    \undefined \def \showISBNx     #1{\unskip}     \fi
\ifx \showISBNxiii \undefined \def \showISBNxiii  #1{\unskip}     \fi
\ifx \showISSN     \undefined \def \showISSN      #1{\unskip}     \fi
\ifx \showLCCN     \undefined \def \showLCCN      #1{\unskip}     \fi
\ifx \shownote     \undefined \def \shownote      #1{#1}          \fi
\ifx \showarticletitle \undefined \def \showarticletitle #1{#1}   \fi
\ifx \showURL      \undefined \def \showURL       {\relax}        \fi
\providecommand\bibfield[2]{#2}
\providecommand\bibinfo[2]{#2}
\providecommand\natexlab[1]{#1}
\providecommand\showeprint[2][]{arXiv:#2}

\bibitem[Baitieva et~al\mbox{.}(2024)]%
        {baitieva2024supervised}
\bibfield{author}{\bibinfo{person}{Aimira Baitieva}, \bibinfo{person}{David Hurych}, \bibinfo{person}{Victor Besnier}, {and} \bibinfo{person}{Olivier Bernard}.} \bibinfo{year}{2024}\natexlab{}.
\newblock \showarticletitle{Supervised Anomaly Detection for Complex Industrial Images}. In \bibinfo{booktitle}{\emph{Proceedings of the IEEE/CVF Conference on Computer Vision and Pattern Recognition}}. \bibinfo{pages}{17754--17762}.
\newblock


\bibitem[Chen et~al\mbox{.}(2017)]%
        {chen2017targeted}
\bibfield{author}{\bibinfo{person}{Xinyun Chen}, \bibinfo{person}{Chang Liu}, \bibinfo{person}{Bo Li}, \bibinfo{person}{Kimberly Lu}, {and} \bibinfo{person}{Dawn Song}.} \bibinfo{year}{2017}\natexlab{}.
\newblock \showarticletitle{Targeted backdoor attacks on deep learning systems using data poisoning}.
\newblock \bibinfo{journal}{\emph{arXiv preprint arXiv:1712.05526}} (\bibinfo{year}{2017}).
\newblock


\bibitem[Cheng and Yuan(2024)]%
        {cheng2024backdoor}
\bibfield{author}{\bibinfo{person}{He Cheng} {and} \bibinfo{person}{Shuhan Yuan}.} \bibinfo{year}{2024}\natexlab{}.
\newblock \showarticletitle{Backdoor Attack Against One-Class Sequential Anomaly Detection Models}. In \bibinfo{booktitle}{\emph{Pacific-Asia Conference on Knowledge Discovery and Data Mining}}. Springer, \bibinfo{pages}{262--274}.
\newblock


\bibitem[Gu et~al\mbox{.}(2017)]%
        {gu2017badnets}
\bibfield{author}{\bibinfo{person}{Tianyu Gu}, \bibinfo{person}{Brendan Dolan-Gavitt}, {and} \bibinfo{person}{Siddharth Garg}.} \bibinfo{year}{2017}\natexlab{}.
\newblock \showarticletitle{Badnets: Identifying vulnerabilities in the machine learning model supply chain}.
\newblock \bibinfo{journal}{\emph{arXiv preprint arXiv:1708.06733}} (\bibinfo{year}{2017}).
\newblock


\bibitem[Gu et~al\mbox{.}(2019)]%
        {gu2019badnets}
\bibfield{author}{\bibinfo{person}{Tianyu Gu}, \bibinfo{person}{Kang Liu}, \bibinfo{person}{Brendan Dolan-Gavitt}, {and} \bibinfo{person}{Siddharth Garg}.} \bibinfo{year}{2019}\natexlab{}.
\newblock \showarticletitle{Badnets: Evaluating backdooring attacks on deep neural networks}.
\newblock \bibinfo{journal}{\emph{IEEE Access}}  \bibinfo{volume}{7} (\bibinfo{year}{2019}), \bibinfo{pages}{47230--47244}.
\newblock


\bibitem[Ionescu et~al\mbox{.}(2019)]%
        {ionescu2019object}
\bibfield{author}{\bibinfo{person}{Radu~Tudor Ionescu}, \bibinfo{person}{Fahad~Shahbaz Khan}, \bibinfo{person}{Mariana-Iuliana Georgescu}, {and} \bibinfo{person}{Ling Shao}.} \bibinfo{year}{2019}\natexlab{}.
\newblock \showarticletitle{Object-centric auto-encoders and dummy anomalies for abnormal event detection in video}. In \bibinfo{booktitle}{\emph{Proceedings of the IEEE/CVF conference on computer vision and pattern recognition}}. \bibinfo{pages}{7842--7851}.
\newblock


\bibitem[Krizhevsky and Hinton(2009)]%
        {krizhevsky2009learning}
\bibfield{author}{\bibinfo{person}{Alex Krizhevsky} {and} \bibinfo{person}{Geoffrey Hinton}.} \bibinfo{year}{2009}\natexlab{}.
\newblock \bibinfo{booktitle}{\emph{Learning multiple layers of features from tiny images}}.
\newblock \bibinfo{type}{{T}echnical {R}eport}. \bibinfo{institution}{Citeseer}.
\newblock


\bibitem[LeCun et~al\mbox{.}(1998)]%
        {lecun1998mnist}
\bibfield{author}{\bibinfo{person}{Yann LeCun}, \bibinfo{person}{Corinna Cortes}, {and} \bibinfo{person}{CJ Burges}.} \bibinfo{year}{1998}\natexlab{}.
\newblock \showarticletitle{The MNIST database of handwritten digits}.
\newblock \bibinfo{journal}{\emph{http://yann.lecun.com/exdb/mnist}} (\bibinfo{year}{1998}).
\newblock


\bibitem[Liu et~al\mbox{.}(2017)]%
        {liu2017trojaning}
\bibfield{author}{\bibinfo{person}{Yingqi Liu}, \bibinfo{person}{Shiqing Ma}, \bibinfo{person}{Yousra Aafer}, \bibinfo{person}{Wen-Chuan Lee}, \bibinfo{person}{Juan Zhai}, \bibinfo{person}{Weihang Wang}, {and} \bibinfo{person}{Xiangyu Zhang}.} \bibinfo{year}{2017}\natexlab{}.
\newblock \showarticletitle{Trojaning attack on neural networks}.
\newblock  (\bibinfo{year}{2017}).
\newblock


\bibitem[Liu et~al\mbox{.}(2020)]%
        {liu2020reflection}
\bibfield{author}{\bibinfo{person}{Yunfei Liu}, \bibinfo{person}{Xingjun Ma}, \bibinfo{person}{James Bailey}, {and} \bibinfo{person}{Feng Lu}.} \bibinfo{year}{2020}\natexlab{}.
\newblock \showarticletitle{Reflection backdoor: A natural backdoor attack on deep neural networks}. In \bibinfo{booktitle}{\emph{Computer Vision--ECCV 2020: 16th European Conference, Glasgow, UK, August 23--28, 2020, Proceedings, Part X 16}}. Springer, \bibinfo{pages}{182--199}.
\newblock


\bibitem[Liu et~al\mbox{.}(2023)]%
        {liu2023generating}
\bibfield{author}{\bibinfo{person}{Zuhao Liu}, \bibinfo{person}{Xiao-Ming Wu}, \bibinfo{person}{Dian Zheng}, \bibinfo{person}{Kun-Yu Lin}, {and} \bibinfo{person}{Wei-Shi Zheng}.} \bibinfo{year}{2023}\natexlab{}.
\newblock \showarticletitle{Generating anomalies for video anomaly detection with prompt-based feature mapping}. In \bibinfo{booktitle}{\emph{Proceedings of the IEEE/CVF conference on computer vision and pattern recognition}}. \bibinfo{pages}{24500--24510}.
\newblock


\bibitem[Nguyen and Tran(2021)]%
        {nguyen2021wanet}
\bibfield{author}{\bibinfo{person}{Anh Nguyen} {and} \bibinfo{person}{Anh Tran}.} \bibinfo{year}{2021}\natexlab{}.
\newblock \showarticletitle{Wanet--imperceptible warping-based backdoor attack}.
\newblock \bibinfo{journal}{\emph{arXiv preprint arXiv:2102.10369}} (\bibinfo{year}{2021}).
\newblock


\bibitem[Pang et~al\mbox{.}(2021)]%
        {pang2021deep}
\bibfield{author}{\bibinfo{person}{Guansong Pang}, \bibinfo{person}{Chunhua Shen}, \bibinfo{person}{Longbing Cao}, {and} \bibinfo{person}{Anton Van~Den Hengel}.} \bibinfo{year}{2021}\natexlab{}.
\newblock \showarticletitle{Deep learning for anomaly detection: A review}.
\newblock \bibinfo{journal}{\emph{ACM computing surveys (CSUR)}} \bibinfo{volume}{54}, \bibinfo{number}{2} (\bibinfo{year}{2021}), \bibinfo{pages}{1--38}.
\newblock


\bibitem[Roth et~al\mbox{.}(2022)]%
        {roth2022towards}
\bibfield{author}{\bibinfo{person}{Karsten Roth}, \bibinfo{person}{Latha Pemula}, \bibinfo{person}{Joaquin Zepeda}, \bibinfo{person}{Bernhard Sch{\"o}lkopf}, \bibinfo{person}{Thomas Brox}, {and} \bibinfo{person}{Peter Gehler}.} \bibinfo{year}{2022}\natexlab{}.
\newblock \showarticletitle{Towards total recall in industrial anomaly detection}. In \bibinfo{booktitle}{\emph{Proceedings of the IEEE/CVF conference on computer vision and pattern recognition}}. \bibinfo{pages}{14318--14328}.
\newblock


\bibitem[Ruff et~al\mbox{.}(2021)]%
        {ruff2021unifying}
\bibfield{author}{\bibinfo{person}{Lukas Ruff}, \bibinfo{person}{Jacob~R Kauffmann}, \bibinfo{person}{Robert~A Vandermeulen}, \bibinfo{person}{Gr{\'e}goire Montavon}, \bibinfo{person}{Wojciech Samek}, \bibinfo{person}{Marius Kloft}, \bibinfo{person}{Thomas~G Dietterich}, {and} \bibinfo{person}{Klaus-Robert M{\"u}ller}.} \bibinfo{year}{2021}\natexlab{}.
\newblock \showarticletitle{A unifying review of deep and shallow anomaly detection}.
\newblock \bibinfo{journal}{\emph{Proc. IEEE}} \bibinfo{volume}{109}, \bibinfo{number}{5} (\bibinfo{year}{2021}), \bibinfo{pages}{756--795}.
\newblock


\bibitem[Ruff et~al\mbox{.}(2018)]%
        {ruff2018deep}
\bibfield{author}{\bibinfo{person}{Lukas Ruff}, \bibinfo{person}{Robert Vandermeulen}, \bibinfo{person}{Nico Goernitz}, \bibinfo{person}{Lucas Deecke}, \bibinfo{person}{Shoaib~Ahmed Siddiqui}, \bibinfo{person}{Alexander Binder}, \bibinfo{person}{Emmanuel M{\"u}ller}, {and} \bibinfo{person}{Marius Kloft}.} \bibinfo{year}{2018}\natexlab{}.
\newblock \showarticletitle{Deep one-class classification}. In \bibinfo{booktitle}{\emph{International conference on machine learning}}. PMLR, \bibinfo{pages}{4393--4402}.
\newblock


\bibitem[Ruff et~al\mbox{.}(2020)]%
        {ruff2020deep}
\bibfield{author}{\bibinfo{person}{Lukas Ruff}, \bibinfo{person}{Robert~A. Vandermeulen}, \bibinfo{person}{Nico G{\"o}rnitz}, \bibinfo{person}{Alexander Binder}, \bibinfo{person}{Emmanuel M{\"u}ller}, \bibinfo{person}{Klaus-Robert M{\"u}ller}, {and} \bibinfo{person}{Marius Kloft}.} \bibinfo{year}{2020}\natexlab{}.
\newblock \showarticletitle{Deep Semi-Supervised Anomaly Detection}. In \bibinfo{booktitle}{\emph{International Conference on Learning Representations}}.
\newblock


\bibitem[Turner et~al\mbox{.}(2018)]%
        {turner2018clean}
\bibfield{author}{\bibinfo{person}{Alexander Turner}, \bibinfo{person}{Dimitris Tsipras}, {and} \bibinfo{person}{Aleksander Madry}.} \bibinfo{year}{2018}\natexlab{}.
\newblock \showarticletitle{Clean-label backdoor attacks}.
\newblock  (\bibinfo{year}{2018}).
\newblock


\bibitem[Wolleb et~al\mbox{.}(2022)]%
        {wolleb2022diffusion}
\bibfield{author}{\bibinfo{person}{Julia Wolleb}, \bibinfo{person}{Florentin Bieder}, \bibinfo{person}{Robin Sandk{\"u}hler}, {and} \bibinfo{person}{Philippe~C Cattin}.} \bibinfo{year}{2022}\natexlab{}.
\newblock \showarticletitle{Diffusion models for medical anomaly detection}. In \bibinfo{booktitle}{\emph{International Conference on Medical image computing and computer-assisted intervention}}. Springer, \bibinfo{pages}{35--45}.
\newblock


\bibitem[Xiao et~al\mbox{.}(2017)]%
        {xiao2017fashion}
\bibfield{author}{\bibinfo{person}{Han Xiao}, \bibinfo{person}{Kashif Rasul}, {and} \bibinfo{person}{Roland Vollgraf}.} \bibinfo{year}{2017}\natexlab{}.
\newblock \showarticletitle{Fashion-MNIST: A novel image dataset for benchmarking machine learning algorithms}.
\newblock \bibinfo{journal}{\emph{arXiv preprint arXiv:1708.07747}} (\bibinfo{year}{2017}).
\newblock


\bibitem[Zhang et~al\mbox{.}(2023)]%
        {zhang2023model}
\bibfield{author}{\bibinfo{person}{Yinghao Zhang}, \bibinfo{person}{Donghuan Lu}, \bibinfo{person}{Munan Ning}, \bibinfo{person}{Liansheng Wang}, \bibinfo{person}{Dong Wei}, {and} \bibinfo{person}{Yefeng Zheng}.} \bibinfo{year}{2023}\natexlab{}.
\newblock \showarticletitle{A Model-Agnostic Framework for Universal Anomaly Detection of Multi-organ and Multi-modal Images}. In \bibinfo{booktitle}{\emph{International Conference on Medical Image Computing and Computer-Assisted Intervention}}. Springer, \bibinfo{pages}{232--241}.
\newblock


\end{thebibliography}

\end{document}